\documentclass[sn-mathphys-num,iicol]{sn-jnl}% Math and Physical Sciences Numbered Reference Style 
\usepackage{multirow}
\usepackage{graphicx}%
\usepackage{multirow}%
\usepackage{amsmath,amssymb,amsfonts}%
\usepackage{amsthm}%
\usepackage{mathrsfs}%
\usepackage[title]{appendix}%
\usepackage{xcolor}%
\usepackage{textcomp}%
\usepackage{manyfoot}%
\usepackage{booktabs}%
\usepackage{algorithm}%
\usepackage{algorithmicx}%
\usepackage{algpseudocode}%
\usepackage{arydshln}
\usepackage{listings}%
\usepackage{color}
\usepackage{colortbl}
\usepackage{subfloat}
\usepackage{subfig}
\usepackage{amsthm}
\usepackage{bm}
\usepackage{verbatim}
\usepackage{graphicx}
\usepackage[T1]{fontenc}
\usepackage{stfloats}
\usepackage{array}
\usepackage{setspace}
\usepackage{multirow} % for cmd 'multirow', 'multicolumn'
\usepackage{setspace}
\usepackage{natbib}
\usepackage{times}
\usepackage{epsfig}
\usepackage{url}
\usepackage{rotate}
\setcitestyle{authoryear,open={(},close={)}} % cite style
\usepackage{hyperref}
\usepackage{color}

\theoremstyle{thmstyleone}%
\theoremstyle{thmstyletwo}%

\theoremstyle{thmstylethree}%

\begin{document}

%\title{\textcolor{red}{DUPE: Downstream-driven Underwater Perceptual Enhancement with Heuristic Invertible Network}}
\title{HUPE: Heuristic Underwater Perceptual Enhancement with Semantic Collaborative Learning}

\author[1]{\fnm{Zengxi} \sur{Zhang}}\email{cyouzoukyuu@gmail.com}

\author[2]{\fnm{Zhiying} \sur{Jiang}}\email{zyjiang0630@gmail.com}

\author[1]{\fnm{Long} \sur{Ma}}\email{malone94319@gmail.com}
\author[1]{\fnm{Jinyuan} \sur{Liu}}\email{atlantis918@hotmail.com}
\author[1]{\fnm{Xin} \sur{Fan}}\email{xin.fan@dlut.edu.cn}
\author*[1,3]{\fnm{Risheng} \sur{Liu}}\email{rsliu@dlut.edu.cn}

\affil[1]{\orgdiv{School of Software Engineering}, \orgname{Dalian University of Technology}, \orgaddress{ \city{Dalian}, \postcode{116024}, \country{China}}}
\affil[2]{\orgdiv{College of Information Science and Technology}, \orgname{Dalian Maritime University}, \orgaddress{ \city{Dalian}, \postcode{116026}, \country{China}}}
\affil[3]{\orgdiv{Pazhou Laboratory (Huangpu)}, \orgaddress{ \city{Guangzhou}, \postcode{510555}, \country{China}}}

%% ================================== %%
%%  Sample for unstructured abstract  %%
%% ================================== %%

\abstract{
Underwater images are often affected by light refraction and absorption, reducing visibility and interfering with subsequent applications. 
Existing underwater image enhancement methods primarily focus on improving visual quality while overlooking practical implications.
To strike a balance between visual quality and application, we propose a heuristic invertible network for underwater perception enhancement, dubbed HUPE, which enhances visual quality and demonstrates flexibility in handling other downstream tasks. 
Specifically, we introduced an information-preserving reversible transformation with embedded Fourier transform to establish a bidirectional mapping between underwater images and their clear images. Additionally, a heuristic prior is incorporated into the enhancement process to better capture scene information. 
To further bridge the feature gap between vision-based enhancement images and application-oriented images, a semantic collaborative learning module is applied in the joint optimization process of the visual enhancement task and the downstream task, which guides the proposed enhancement model to extract more task-oriented semantic features while obtaining visually pleasing images.
Extensive experiments, both quantitative and qualitative, demonstrate the superiority of our HUPE over state-of-the-art methods. The source code is available at \href{https://github.com/ZengxiZhang/HUPE}{https://github.com/ZengxiZhang/HUPE}.
}

\keywords{Underwater image enhancement, deep learning, color correlation, object detection}


\maketitle

\section{Introduction}\label{sec1}

As underwater robotics forge ahead, there is a surging market appetite for applications like underwater tracking~\citep{walther2004detection,cai2023semi}, animal identification~\citep{hsiao2014real,hughes2017automated,mcever2023context}, and marine grasping~\citep{cai2020grasping}. However, in contrast to their terrestrial counterparts, water serving as a medium for light transmission, induces refraction and absorption effects on light. Images captured underwater often suffer from low contrast and color artifacts, which can significantly impact the accuracy of subsequent perception applications. 
Therefore, underwater image enhancement plays a crucial role in underwater applications, which mitigate the impact of underwater imaging degradation on perceptual tasks.

Existing traditional methods utilize degradation models-based image enhancement as a preprocessing stage, providing the enhanced images as potential inputs to subsequent perception applications. However, these fixed models constrain the scene representation of complex environments. With the advancement of deep learning for low-level visual tasks, the trend toward end-to-end networks for underwater image enhancement is becoming increasingly evident. However, current methods primarily emphasize visual improvements, neglecting the restoration of the perceptual quality, thus offering limited benefits for downstream tasks. A series of methods~\citep{lee2020self,chen2020deep,chi2021test} have been proposed for other low-level enhancement tasks to bolster the adaptability of enhanced models to downstream applications. However, a common drawback with existing technologies is their tendency to merely concatenate the enhancement network with the downstream task network, thereby neglecting the unified representation in the feature space, which ultimately limits their effectiveness.

To balance visual enhancement and subsequent applications, in this paper, we propose a heuristic invertible network to achieve underwater perceptual enhancement. 
Specifically, we first introduce the invertible network to achieve the enhancement of the underwater image by constructing a reversible translation of the underwater image and its clear counterpart.
The forward translation process functions as an enhancement process to describe the manifold structure of in-air images, while the backward translation process acts as a constraint, effectively mitigating artifacts and preventing information loss.
Additionally, the heuristic prior information is integrated into the translation process, augmenting its capability to better represent various complex underwater environments.

Due to the considerable disparity in vision enhancement and semantic excavation, simple joint learning with a cascade of upstream and downstream tasks fails to effectively align visual-quality enhancement with downstream tasks. 
To address this issue, a semantic collaborative learning module is further introduced during the training process.  
The gap between visual appearance and high-level semantic perception features is mitigated by integrating a feature collaboration module between the enhancement network and the downstream task network.
Consequently, the enhancement network not only achieves visual enhancement but also further extracts semantic features from the image, thereby realizing perceptual enhancement.
In summary, contributions of this work can be concluded as follows:
\begin{itemize}
	\item We apply the information-preserving invertible network to the underwater image enhancement task, enabling an invertible translation between the degraded images and their clear counterparts.
	\item We integrate the heuristic prior into the data-driven mapping process, enhancing the resilience of the method in real-world underwater scenarios by augmenting the interpretability of the entire framework.
	\item We propose a semantic collaborative learning module, which guides the proposed network to generate visually pleasing images while simultaneously extracting high-level semantic information through a feature-level collaborative learning.
	\item Both qualitative and quantitative analyses affirm that the proposed HUPE effectively restores the original reflection of the intrinsic scene, making it more adept for perceptual tasks.
\end{itemize}
The preliminary version of this paper is referred to as WaterFlow~\citep{zhang2023waterflow}. In contrast to previous methods, the main improvements of this works can be summarized as follows:
\begin{itemize}
	\item We conduct the Frequency-Aware Affine Coupling Layer, 
	which helps the network better characterize the internal relationship 
	in both the spatial domain and frequency domain by introducing Fourier transform in the network.
	\item We propose the Feature Collaboration Module, which enables the enhanced network to further learn to extract semantic features of underwater images while generating visually pleasing images.
	\item Instead of solely focusing on extracting semantic information for underwater object detection, we further extend the application to underwater semantic segmentation tasks.
\end{itemize}

\section{RELATED WORK}
\subsection{Underwater Perceptual Enhancement}
Different from conventional visual enhancement methods, in this paper, the proposed method achieves perceptual enhancement of underwater images, which includes the visual quality enhancement and the excavation of intrinsic semantic information.

\textbf{Underwater Image Enhancement:}
As a crucial step in underwater visual perception, various underwater image enhancement methods have been introduced in recent years. These enhancement methods can be broadly categorized into visually guided  methods, physical model-based methods, and deep learning-based methods.
Vision-guided methods usually restore the degradation of light underwater by adjusting the pixel distribution of the different color channels.
\cite{iqbal2010enhancing} utilized the saturation and intensity information of the HSI color model to correct the contrast of images while balancing color values.  
\cite{hitam2013mixture} conducted adaptive histogram equalization separately on the RGB and HSV color channels, followed by fusion using the Euclidean norm to enhance the visual quality of the underwater image.
Taking into account the varying degrees of light attenuation in underwater images, \cite{ghani2015underwater} extended the proportions of the primary color channels downward distribution and increased the proportions of the secondary color channels  based on the Rayleigh distribution. 
\cite{li2016single} combined the widely used Grey World, White Patch, and Histogram Equalization to correct the contrast of both the RGB and HSI channels, aiming to restore the authentic colors in underwater images.
While these methods can mitigate the fading of light on various color channels, the absence of guidance from physical models makes it more prone to introduce artificial colors into underwater images.

Physical model-based methods usually introduce physical imaging model~\cite{chiang2011underwater,drews2013formationmodel,ye2022perceiving} as priors to enhance the robustness of the method.
\cite{drews2013formationmodel} introduced a dark channel prior calculated based on image characteristics obtained in outdoor natural scenes to estimate the light transmission process underwater.
Through background light region detection and the incorporation of underwater optical characteristics, \cite{li2017hybrid} introduced a dedicated global background light estimation algorithm tailored specifically for underwater images. 
\cite{wang2017single} introduced a  pixel distribution-depended non-local prior to estimate the transmission and attenuation factors of underwater images, enabling compensation for transmission effects.
\cite{peng2018generalization} estimated ambient light based on depth-dependent color changes, and incorporate adaptive color correction within the image formation model to eliminate color casts and restore contrast.

With the development of deep learning, many low-level image enhancement methods based on deep learning~\citep{fan2022multiscale,liu2022attention,jiang2022topal,liu2024coconet,zhang2024synergistic,wei2024learning,jiang2025drnet} have introduced in recent years.
DLIFM~\citep{chen2021dlifm} used the convolutional neural network to simulate backscatter estimation and transmission.
Ucolor~\citep{li2021ucolor} separately encoded features in multi-color spaces, and combined the attention mechanism to adaptively integrate the encoded features.
SIBM~\citep{mu2022structure} combined knowledge patterns such as semantics, gradients, and pixels to hierarchically enhance underwater images, and introduced an optimization scheme for hyperparameters to fuse the above information.
TACL~\citep{liu2022tacl} introduced unsupervised training methods into coupled twin mapping, which alleviated the requirement of paired training data while retaining more informative features.
CECF~\citep{cong2024underwater} achieves mapping of degraded images and clear images through color code decomposition combined with content-invariant learning.
%Jiang~\emph{et al.}~\cite{jiang2022topal} employed a multi-scale dense network to enhance image contrast and introduce an aesthetic rendering module for color correction. 
SemiUIR~\citep{huang2023contrastive} proposed a semi-supervised underwater enhancement network utilizing a mean-teacher strategy and employed contrastive regularization to prevent overfitting. 
\cite{hclrnet2024zhou} applied the constructed negative samples to the hybrid contrastive learning regularization strategy to enhance the magical generalization ability.

\textbf{Underwater Scene Perception:}
With the development of underwater robots, more and more works are focused on the perception of underwater objects, e.g., detection and segmentation.
For object detection tasks, \cite{yeh2021lightweight} achieve lightweight object detection by converting the color of underwater images into corresponding grayscale images.
\cite{zeng2021underwater} introduced the adversarial occlusion network into the standard Fatser R-CNN detection network, and achieved more robust underwater object detection through balanced training of the dual networks.
As for semantic segmentation tasks,
\cite{2020SUIM} first proposed a fully convolutional deep residual model for underwater images to achieve good performance with limited computing resources.
\cite{li2021marineseg} proposed a cascaded decoder network consisting of multiple interactive feature enhancement modules to achieve segmentation of marine animals from complex environments.

\begin{figure*}[tp]
	\centering
	\setlength{\tabcolsep}{1pt}
	\includegraphics[width=\textwidth]{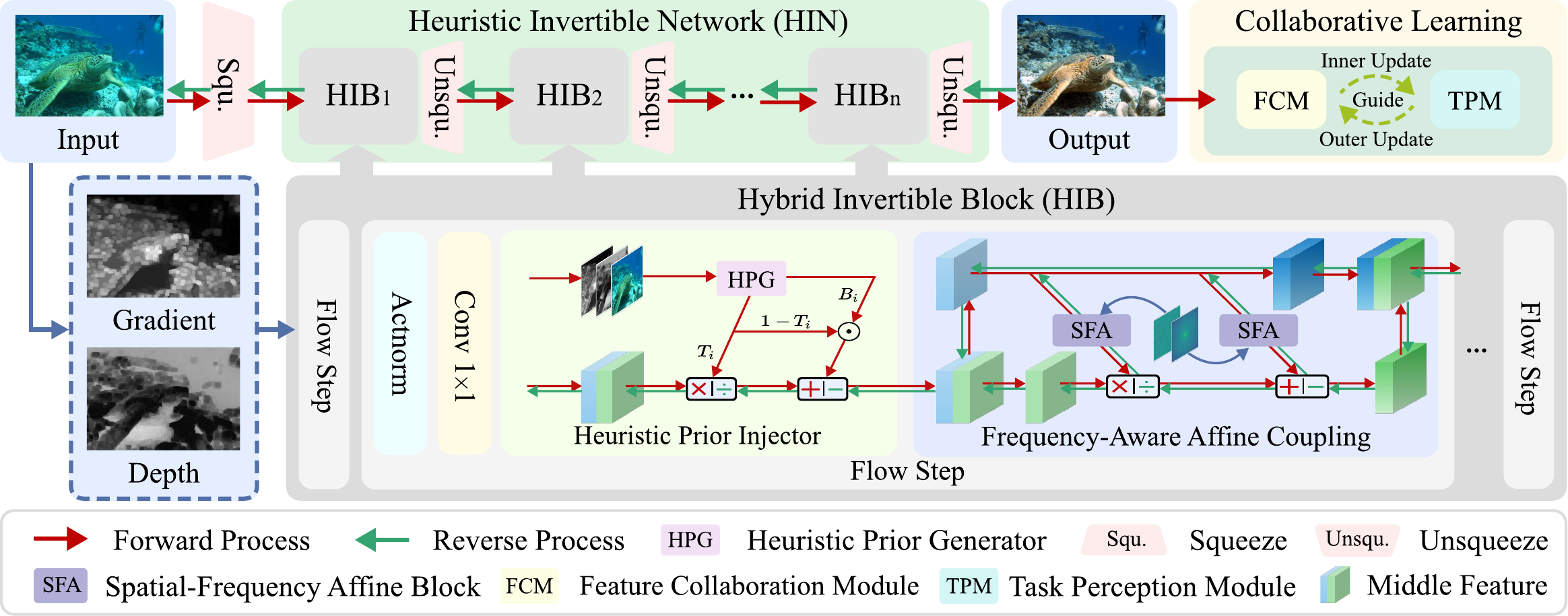}
	\caption{Workflow of the proposed HUPE. Physical model-related depth and gradient information are embedded into the reversible mapping of degraded images and clear counterparts to assist the network in generating credible enhanced results. A semantic collaborative learning module is introduced to assist the enhancement network to maximize the retention and extraction of the semantic structure of the image.}
	\label{fig:workflow}
\end{figure*}

\subsection{Collaborative Learning}
With the development of computer vision, many works have proposed joint optimization methods to simultaneously achieve low-level visual restoration and downstream perception tasks~\citep{jiang2023bilevel,liu2023value,liu2023hierarchical,wu2024hybrid,liu2024learning}. 
\cite{you2019unsupervised} incorporated image restoration into the anomaly detection task, and further detected abnormal lesions through the difference between the original and restored images.
\cite{chen2020deep} proposed a joint optimization framework to achieve image restoration and image recognition through cascading a unified common feature extraction layer with an optional restoration layer and classification layer.
Considering that existing restoration networks show limit effection on improving the performance of detectors, \cite{sun2022rethinking} proposed an adversarial optimization model based on ADAM attack to improve the performance of object detection during image restoration.
%\cite{liu2022learning} established a general learning framework inspired by Retinex that achieves low-light image enhancement while flexibly handling downstream vision applications.
\cite{zhao2023metafusion} embedded meta-features into the image fusion task to further improve the performance of fused images on the object detection task.
\cite{liu2024task} proposed a deep model based on nested learning to address the problem of image fusion in complex scenes through an implicit search scheme.
The proposed method attempts to reduce the difference between visual features and semantic features, thereby achieving collaborative perceptual enhancement from underwater images.

\section{The Proposed Method}
Traditional learning-based enhancement networks, due to their irreversible characteristic, tend to result in the loss of crucial details, color fidelity, and semantic structures during the enhancement process, consequently undermining the accuracy and reliability of subsequent perception tasks. 
To address this issue, we first propose the Heuristic Invertible Network~(HIN), which employs reversible translation between underwater and clear images to preserve information integrity during enhancement.
The Frequency-Aware Affine Coupling Module is embedded in the translation, which synchronously characterizes the intrinsic relationship in both frequency and spatial domains between underwater images and their clear counterparts.
Considering that the data-driven network often overlooks an understanding of underwater scenes, resulting in less adaptability to varied and complex environments, we additionally encode physical model-related depth and gradient information as heuristic priors into the invertible network. This approach enhances the robustness of the enhancement network across diverse underwater conditions while minimizing reliance on extensive training data.

Current underwater enhancement methods usually focus solely on vision-based enhancement, neglecting the preservation and augmentation of perceptual features. 
A question worth pondering is whether there is a possibility to reduce the gap in feature space between visual enhancement tasks and downstream applications.
To overcome this problem, the Semcntic Collaborative Learning Module~(SCL) is introduced to further enhance the adaptability of the enhanced images to downstream perceptual tasks. By embedding meta-features between the enhancement network and subsquent perception network, the proposed HUPE can further implicitly excavate semantic information from the image, so that it can be better applied to subsequent perception tasks. 
The following section provides a detailed introduction to the structure and the training strategy of the method.

\subsection{Hybrid Invertible Block}

Fig.~\ref{fig:workflow} demonstrates the workflow of the HUPE.
The Hybrid Invertible Block (HIB) is the core component of the heuristic invertible network, responsible for facilitating reversible translation while embedding heuristic prior information into the network to bolster its enhancement robustness. In the forward process, the underwater image is transformed into the enhanced image via the invertible network composed of multiple invertible blocks. Conversely, in the reverse process, the clear image undergoes inverse operations of the invertible network to reconstruct the degraded image.

The Hybrid Invertible Block consists of Actnorm~\citep{kingma2018glow}, Invertible Conv 1$\times$1~\citep{kingma2018glow}, Heuristic Prior Injector, Frequency-Aware Affine Coupling and Unsqueeze/squeeze operation.

\textbf{Actnorm}: It employs per-channel scale and bias parameters for an affine transformation, initially normalized for zero mean and unit variance based on the first data batch. 

\textbf{Conv $\textbf{1} \times \textbf{1}$}: It is an extension of permutation operation, which alters the number of channels in the input by using a 1x1 convolutional kernel while preserving its spatial dimensions. This operation allows the model to perform linear combinations across different feature channels.

\textbf{Heuristic Prior Injector}: 
It is introduced to further improve the network's ability to learn underwater scenes through guidance from the physical imaging parameters according to the simple deformed underwater image imaging model~\citep{chiang2011underwater,drews2013formationmodel}, which can be expressed as:
\begin{equation}
	J^c(x) = \frac{1}{t(x)}I^c(x) + \frac{1}{t(x)}B^c(t(x)-1), c \in\{r, g, b\},
	\label{eq:imaginginverse}
\end{equation}
where $J$ denotes the enhanced image at location $x$, $I$ denote the underwater image captured by sensors.~$c$ represents the color space of red, green and blue channel.~$B$ represents the ambient light. $t$ indicates and the medium transmission coefficient, it can also be expressed as $t(x)=e^{\beta d(x)}$ with the attenuation coefficient $\beta$ and the scene depth $d(x)$ according to the Beer-Lambert law~\citep{bouguer1729essai}.

Considering that the gradient map assists in correcting the image's blurriness caused by light scattering underwater by estimating the effect of small impurities on light scattering. Additionally, the depth map provides crucial information for estimating atmospheric light and light attenuation underwater. We previously estimate both the gradient map $I_g$ and depth map $I_d$ through the generalization of the dark channel prior~\citep{peng2018generalization} as the heuristic information.

Subsequently, we concatenate them with the degradation image $I_u$ and feed them into the Heuristic Prior guided Encoder, which gradually estimates underwater imaging parameters by encoding the color, gradient, and depth information of the degraded image. The encoding process can be represented as:
\begin{equation}
	B_i,T_i =\mathcal{S}\left(\mathcal{H}_i\left(\mathcal{C}\left(I_u,I_g,I_d\right)\right)\right),
\end{equation}
where~$\mathcal{H}$ represent the Heuristic Prior guided Encoder, whose detailed architecture is illustrated in the previous version~\citep{zhang2023waterflow}. $\mathcal{C}$ and $\mathcal{S}$ respectively denote the channel-wise concatenate and split operation .~$T$ is the reciprocal of $t$, from the Beer-Lambert law~\citep{bouguer1729essai}. After estimating $T$ and $B$, we insert them into the HIB as heuristic information. 
%The invertible formulation can be demonstrated as:
%\begin{equation}
%	\begin{aligned}
%		\mathbf{u}_{i+1} =&~T_i \odot \mathbf{u}_i + B_i\odot(1-T_i),\\
%		\mathbf{u}_{i} =&\left(\mathbf{u}_{i+1} - B_i\odot(1-T_i)\right) / T_i,\\
%	\end{aligned}
%\end{equation}
%63 167 171 244 248 470
%where $\odot$ represents dot product, $\mathbf{u}_i$ denotes the middle feature of the $i$-th flow steps. 
After embedding the heuristic prior into the invertible network, the proposed method can better characterize the intrinsic relationship between the underwater domain and in-air domain with more generalization.
\begin{figure}[tp]
	\centering
	\setlength{\tabcolsep}{1pt}
	\includegraphics[width=0.48\textwidth]{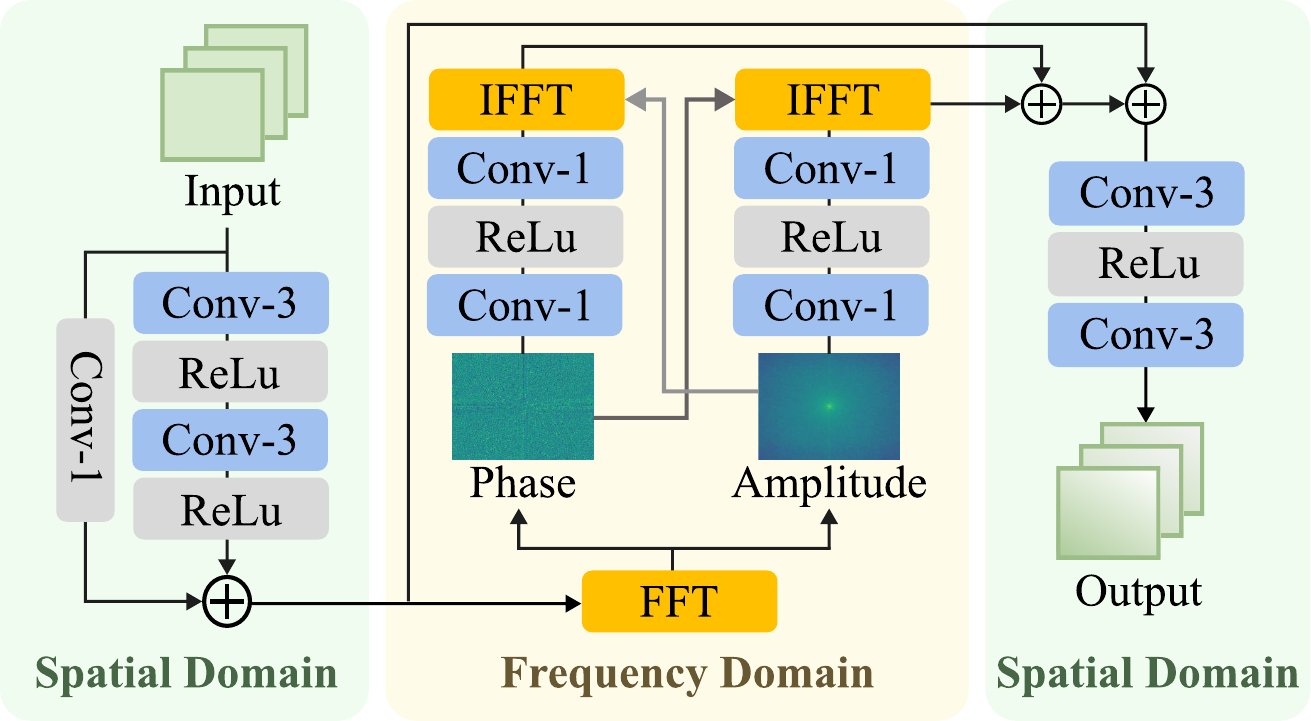}
	%	\vspace{-5pt}
	\caption{Workflow of the proposed Spatial-Frequency Affine Block.}
	\label{fig:SFA}
\end{figure}

\textbf{Frequency-Aware Affine Coupling}: It is proposed to enhance the transformation capability of individual flow steps from input to output. The specific operations will be described in Section~\ref{FAAC}.

\textbf{Unsqueese/Squeeze}: It can achieve reversible feature expansion and compression. Unsqueeze is designed to convert the image from $4 c \times \frac{h}{2} \times \frac{w}{2}$ to $c \times h \times w$. Squeeze is the reverse operation of squeeze. 
In the forward process, we first convert the input image from $c \times h \times w$ to $2^N c \times \frac{h}{2^N} \times \frac{w}{2^N}$, where $N$ represents the number of Hybrid Invertible Block~(HIB). and then gradually convert it back to $c \times h \times w$ through the HIBs. The reverse process is the opposite of the forward process.
%The operations of the forward process and the reverse process are opposite to ensure the reversibility of the process.
Through the integration of these operations, the proposed Heuristic Invertible Network can achieve precise mapping between the underwater image and its clear counterpart.

\subsection{Frequency-Aware Affine Coupling}\label{FAAC}
The Affine Coupling Layer is exploited to further express information during the conversion. Different from the previous affine coupling layer~\citep{li2021dehazeflow}, which have limited exploration of frequency domain information, we apply Fourier transform~\citep{shen2023mutual,yao2024sfle} $\mathcal{F}$ to convert the input into amplitude and phase from spatial domain. The formulation can be expressed as: 
\begin{equation}
	\begin{aligned}
		\mathbf{u}^1_{i+1}=&~\mathbf{u}^1_i, \\
		\mathbf{u}^2_{i+1}=&~\phi_i\left(\mathbf{u}^1_i,\mathcal{F}\right)\odot \mathbf{u}^2_i
		+\phi_i\left(\mathbf{u}^1_i, \mathcal{F}\right),
	\end{aligned}
\end{equation}
where  $\mathbf{u}_{i+1}= \mathcal{C}(\mathbf{u}_{i+1}^1,\mathbf{u}_{i+1}^2)$, in which~$\mathcal{C}$ represents the concatenation of features.~$i \in \{1,...,n-1\}$, which means the index of flow steps. 
The Fourier transform $\mathcal{F}$ can be formulated as follows:
\begin{equation}
	\mathcal{F}\left(\mathbf{x}\right)(i, j)= \sum_{h=0}^{H-1} \sum_{w=0}^{W-1} \mathbf{x}(h, w) e^{-k 2 \pi\left(\frac{h}{H} i+\frac{w}{W} j\right)} .
\end{equation}
The frequency domain $\mathcal{F}(x)=\mathcal{R}(x)+k \mathcal{I}(x)$, where $\mathcal{I}$ and $\mathcal{R}$ denote the imaginary and real components. $\phi_i$ denotes the Spatial-Frequency Affine Block~(SFA).  
The workflow of the SFA is illustrated in Fig.~\ref{fig:SFA}, which contains spatial domain and frequency domain. We first process the spatial domain information through the residual block, and then obtain the phase spectrum $\mathcal{P}(\mathbf{x})$ and amplitude spectrum $\mathcal{A}(\mathbf{x})$ through fast Fourier transform~\citep{brigham1967fast}~(FFT) from the input $\mathbf{x}$, which can be described as: 

\begin{equation}
	\begin{aligned}
		& \mathcal{A}\left(\mathbf{x}\right)(i, j)=\sqrt{\mathcal{R}^2\left(\mathbf{x}\right)(i, j)+\mathcal{I}^2\left(\mathbf{x}\right)(i, j)}, \\
		& \mathcal{P}\left(\mathbf{x}\right)(i, j)=\arctan \left[\frac{\mathcal{I}\left(\mathbf{x}\right)(i, j)}{\mathcal{R}\left(\mathbf{x}\right)(i, j)}\right],
	\end{aligned}
\end{equation}
According to the Fourier theory~\citep{xu2021fourier}, the phase component $\mathcal{P}$ conveys the semantic information of images, while the amplitude component $\mathcal{A}$ displays the style information in the frequency domain.
In order for the network to adaptively learn phase features and amplitude features respectively, we feed the two obtained spectrums into the convolution layer and inversely map it with the other original spectrum to converse back to the spatial domain. Finally, we combine the features of the spatial domains with three convolution layers to get the output feature.

%\subsection{Meta-feature Guided Optimization Module}
\subsection{Semantic Collaborative Learning Module}
\begin{figure}[tp]
	\centering
	\setlength{\tabcolsep}{1pt}
	\includegraphics[width=0.48\textwidth]{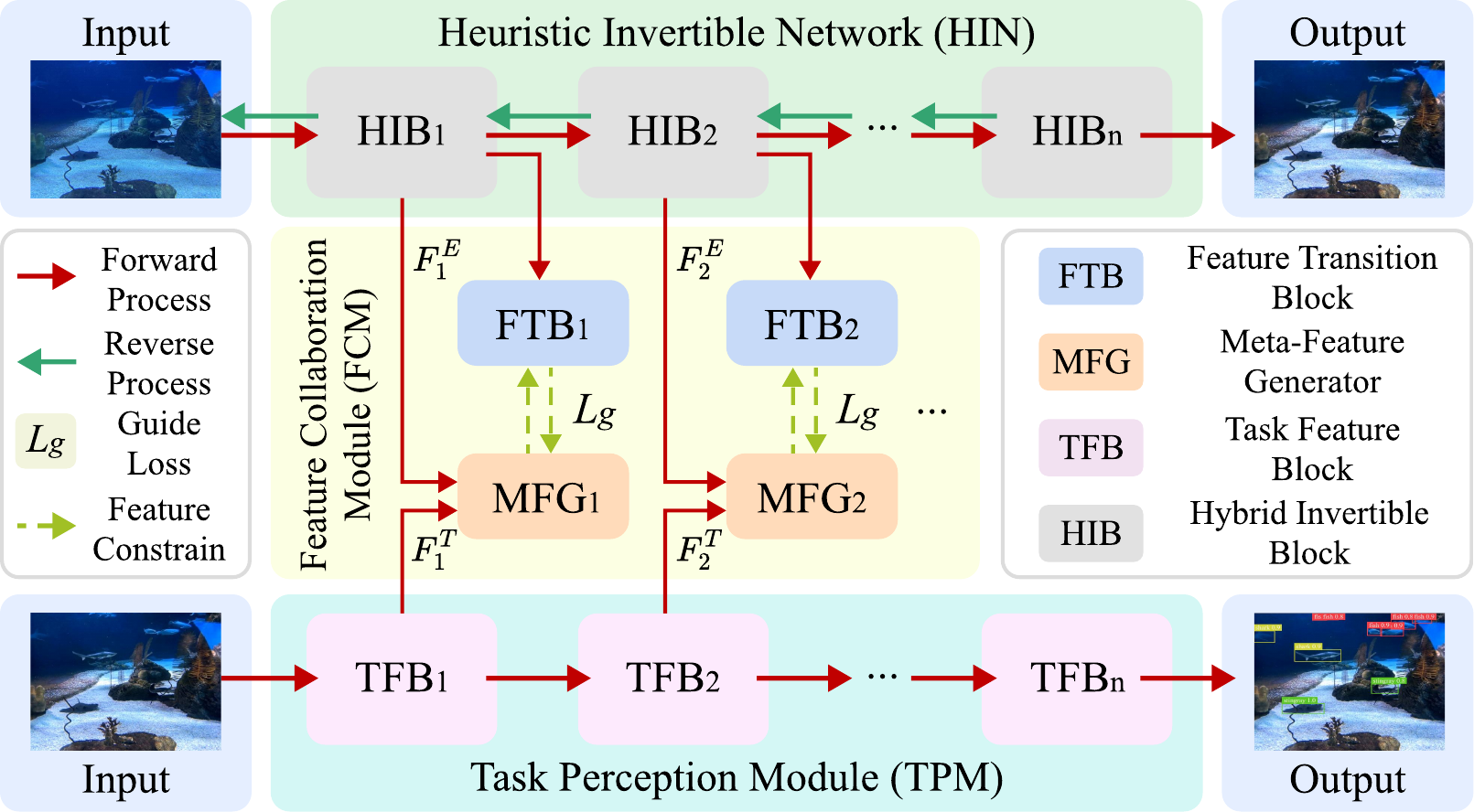}
	\caption{Workflow of the proposed Semantic Collaborative Learning Module.}
	\label{fig:TPM}
\end{figure}

The proposed learning module is shown in Fig.~\ref{fig:TPM}, which includes the proposed enhancement network, the Task Perception Module~(TPM) and the Feature Collaboration Module~(FCM). FCM includes the Meta-feature Generator~(MFG) and the Feature Transformation Block~(FTB). MFG generates meta-features $F^{MFG}_{i}$ from task-aware features $F^{T}_{i}$ and the enhanced features $F^{E}_{i}$ according to the ability for extracting semantic feature of the enhancement network. The formulation can be expressed as $F^{MFG}_{i}=MFG\left(F^{T}_{i}, F^{E}_{i}\right)$, where $i$ is the intermediate feature index. FTB transfers meta-features $F^{MFG}_{i}$ to enhancement features $F^{E}_{i}$ by generating feature bridges $F^{FTB}_{i}$. The detailed architecture of the MFG and FTB are described in Tab.~\ref{tab:details}.

Specifically, the semantic collaborative learning is divided into an internal update stage and an external update stage. It is worth noting that both the enhancement network and the task perception module were pre-trained by enhancement loss $\mathcal{L}_\mathrm{e}$ and task loss $\mathcal{L}_\mathrm{t}$ respectively before the optimization strategy of FCM. 
In the internal update stage, we first optimize the pre-trained Heuristic Invertible Network~(HIN) individually by the guide loss $\mathcal{L}_\mathrm{g}$.% on the meta-training set. 
Then, we use the preliminary optimized HIN to calculate the enhancement loss $\mathcal{L}_\mathrm{e}$ %on the meta-testing set 
for updating the parameters of MFG and FTB. Here, $\mathcal{L}_\mathrm{e}$ can measure the effect of using the meta features to guide HIN. 
By jointly optimizing FTB and MFG, the guidance effect of meta-features on the enhancement network is enhanced. In the external update stage, we further optimize the enhanced network through $\mathcal{L}_\mathrm{g}$ and $\mathcal{L}_\mathrm{e}$. 
Through the alternating optimization of internal stages and external stages, 
the enhanced network can extract more semantic features while generating visually pleasing images, making it more suitable for subsequent perception tasks. 
The details of the loss function used will be shown in the Section~\ref{Loss}.

\begin{table}[!htb]
	\caption{Detailed architecture of the Feature Transition Block~(FTB) and the Meta-Feature Generator~(MFG). out and k represent the output channel and kernel size respectively.}{
		\begin{tabular}{c|c|c|c}
			\toprule
			$\#$ & \text{Input} & \text{Output} & \text{Operation} \\
			\midrule
			\multicolumn{4}{c}{Feature Transition Block}\\
			\midrule
			1 & $F^{\mathrm{E}}_{\mathrm{i}}$ & $z_1$   & Conv2D(out = 128, k = 3), Relu \\
			3 & $z_1$                         & $z_2$   & Conv2D(out = 256, k = 3), Relu \\
			5 & $z_2$ & $F^{\mathrm{FTB}}_{\mathrm{i}}$ & Conv2D(out = 256, k = 3), Relu \\
			\midrule
			\multicolumn{4}{c}{Meta-Feature Generator}\\
			\midrule
			1 & $F^{\mathrm{T}}_{\mathrm{i}}$ & $z_1$ & Conv2D(out = 512, k = 3), Relu \\
			2 & $z_1$  &  $z_2$  & Conv2D(out = 256, k = 3), Relu \\
			3 & $z_2$  &  $z_3$  & Conv2D(out = 256, k = 3), Relu \\
			4 & $z_3$  &  $z_4$  & Conv2D(out = 256, k = 3), Relu \\
			5 & $F^{E}_{i}$      & $z_1$   & Conv2D(out = 128, k = 3), Relu \\
			6 & $z_1$  &  $z_2$  & Conv2D(out = 256, k = 3), Relu \\
			7 & [$z_2$,$z_4$]    & $z_1$   & Conv2D(out = 64, k = 3), Relu \\
			8 & $z_1$  & $z_2$   & Conv2D(out = 128, k = 3), Relu \\
			9 & $z_2$  & $z_3$   & Conv2D(out = 192, k = 3), Relu \\
			10& $z_3$  & $z_4$   & Conv2D(out = 256, k = 3), Relu \\
			11& $z_4$  & $z_5$   & Conv2D(out = 320, k = 3), Relu \\
			12& $z_5$ & $F^{\mathrm{MFG}}_{\mathrm{i}}$ & Conv2D(out = 256,k = 3), Relu \\
			%			\hline
			\bottomrule
	\end{tabular}}
	\label{tab:details}
\end{table}

\subsection{Loss Function}\label{Loss} 
%Multiple loss constraint are employed in the training process, including 
Guide loss $\mathcal{L}_\mathrm{g}$, enhancement loss $\mathcal{L}_\mathrm{e}$, and task loss constraint $\mathcal{L}_\mathrm{t}$ are employed in the training process. 
Guide loss $\mathcal{L}_\mathrm{g}$ calculate the $\mathcal{L}_\mathrm{2}$ distance of the $F^{\mathrm{MFG}}$ and $F^{\mathrm{FTB}}$, which enables the enhancement network guided by perceptual features to be more adaptable to subsequent perceptual tasks.

As for the $\mathcal{L}_\mathrm{e}$, we apply multiple loss in $\mathcal{L}_\mathrm{e}$ based on past experience~\citep{zhang2023waterflow} to make the enhancement image closer to the in-air image. First, contrastive learning is incorporated as the fidelity term in the enhancement process to incentivize the model to learn discriminative features, ultimately improving its ability to extract semantic information. We use degraded underwater-captured images as negative examples and enhancement reference images sampled from the distribution of terrestrial scenes as positive examples. The contrastive loss function $\mathcal{L}_\mathrm{c}$  can be formulated as follows:
\begin{equation}
	\mathcal{L}_{\mathrm{c}}=\sum_{i=1}^N \rho_i 
	\cdot
	\frac{\left\|\mathcal{VGG}_i(I_r)-\mathcal{VGG}_i\left(G_E(I_u)\right)\right\|_1}
	{\left\|\mathcal{VGG}_i(I_u)-\mathcal{VGG}_i\left(G_E(I_u)\right)\right\|_1},
\end{equation}
where $G_E$ is the forward process of the proposed invertible network. $I_r$ and $I_u$ denote the reference image and the underwater image respectively. We employ the pretrained VGG19~\citep{simonyan2014vgg19} to extract features from the respective images. $i \in \{ 1, 3, 5, 9, 13\}$ denotes the $i$-th layer of VGG19. $\rho_i \in \{\frac{1}{32}, \frac{1}{16}, \frac{1}{8}, \frac{1}{4}, 1\}$ represents the corresponding weight of the $i$-th layer. 

In addition, the frequency loss~\citep{shen2023mutual} is introduced to make the frequency domain of the enhanced image closer to the reference image by applying Fourier transform $\mathcal{F}$. The frequency loss function $\mathcal{L}_\mathrm{s}$ is formulated as follows:
\begin{equation}
	\begin{aligned}
		\mathcal{L}_\mathrm{f}=  \left\|\left(\mathcal{F}\left(G_E(I_u)\right)\right)-\left(\mathcal{F}\left(I_r\right)\right)\right\|_1.
	\end{aligned}
\end{equation}

In practice, imposing constraints on the reverse process can ensure the reversibility of underwater image enhancement, thereby improving the reliability of subsequent perception tasks. The L1 norm is used as a bilateral constraint to bring the forward and reverse outputs of the proposed network closer to the reference image and underwater image. The bilateral loss $\mathcal{L}_{\mathrm{b}}$ is expressed as:
\begin{equation}
	\mathcal{L}_{\mathrm{b}}=\left\|G_E(I_u) - I_r\right\|_2 + \left\|G_E^{-1}(I_r) - I_u\right\|_2.
\end{equation}	
%where $G_E^{-1}$ represents the reverse process of the proposed network $G_E$. 
Therefore, the enhancement loss $\mathcal{L}_\mathrm{e}$ of network training is expressed as follows:		
\begin{equation}
	\mathcal{L}_\mathrm{e}=\lambda_1\mathcal{L}_\mathrm{c}+\lambda_2\mathcal{L}_\mathrm{f}+\lambda_3 \mathcal{L}_\mathrm{b}.
\end{equation}	

Task loss $\mathcal{L}_{\mathrm{t}}$ aims to improve the performance of task-specific networks through constraints on specific perception tasks. In this work, we impose corresponding constraints on the underwater object detection task and semantic segmentation task respectively. As for the detection task, the introduced task depended loss $\mathcal{L}_{\mathrm{t}}^{det}$ can be expressed as 
\begin{equation}
	\mathcal{L}_{\mathrm{t}}^{det} = \mathcal{L}_{\mathrm{cla}}+\mathcal{L}_{\mathrm{loc}},
\end{equation}
where $\mathcal{L}_{\mathrm{loc}}$ represent the localization loss, which is incorporated to reduce the disparity between the ground truth label and the predicted label.
$\mathcal{L}_{\text{cla}}$ denotes the classification loss, which is designed to minimize the positional discrepancy  between the ground truth box and the predicted box. Focal loss~\citep{lin2017focalloss} and GIoU loss~\citep{rezatofighi2019giou} were employed as classification loss and localization loss in this work. As for the segmentation task, common to previous works, $\mathcal{L}_{\mathrm{t}}^{seg}$ is defined as:
\begin{equation}
	\mathcal{L}_{\mathrm{t}}^{seg}=-\sum_{\text {class }} \mathbf{s}^* \log (\mathbf{s})
\end{equation}
where $\mathbf{s}^*$ denotes the segmentation label.

\begin{figure*}[!htb]
	\centering
	\setlength{\abovecaptionskip}{0.cm}
	\setlength{\tabcolsep}{1pt}
	\begin{tabular}{cccccc}
		\includegraphics[width=0.192\textwidth,height=0.099\textheight]{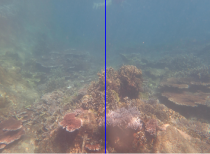}
		&\includegraphics[width=0.192\textwidth,height=0.099\textheight]{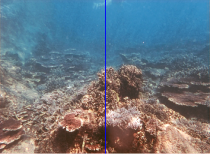}
		&\includegraphics[width=0.192\textwidth,height=0.099\textheight]{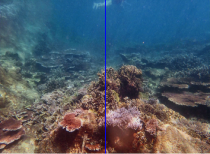}
		&\includegraphics[width=0.192\textwidth,height=0.099\textheight]{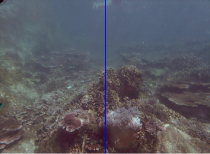}
		&\includegraphics[width=0.192\textwidth,height=0.099\textheight]{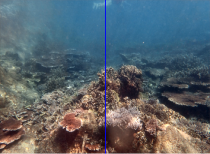}
		\\\includegraphics[width=0.192\textwidth,height=0.093\textheight]{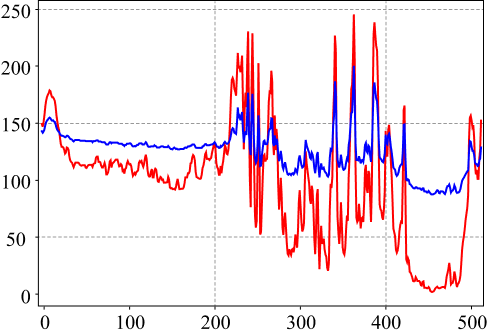}
		&\includegraphics[width=0.192\textwidth,height=0.093\textheight]{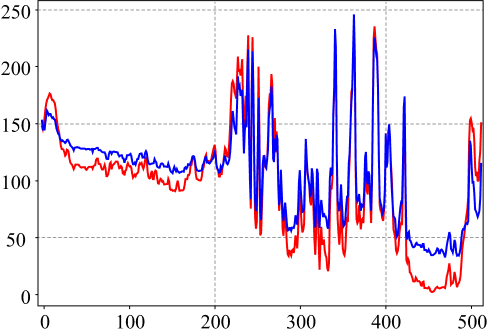}
		&\includegraphics[width=0.192\textwidth,height=0.093\textheight]{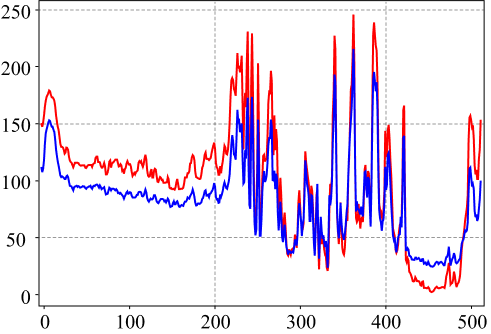}
		&\includegraphics[width=0.192\textwidth,height=0.093\textheight]{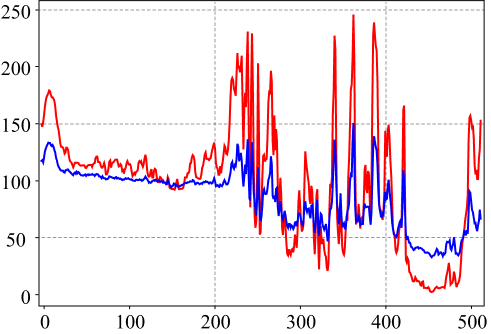}
		&\includegraphics[width=0.192\textwidth,height=0.093\textheight]{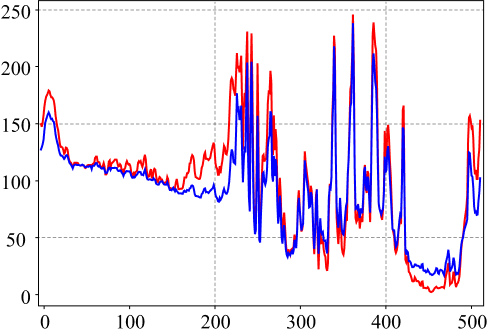}\\
		Input&DLIFM&Ucolor&CWR&TOPAL\\
		\includegraphics[width=0.192\textwidth,height=0.099\textheight]{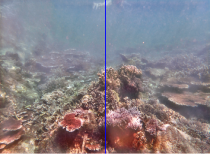}
		&\includegraphics[width=0.192\textwidth,height=0.099\textheight]{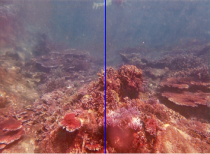}
		&\includegraphics[width=0.192\textwidth,height=0.099\textheight]{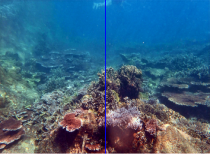}
		&\includegraphics[width=0.192\textwidth,height=0.099\textheight]{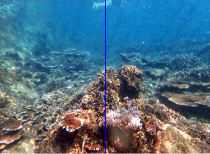}
		&\includegraphics[width=0.192\textwidth,height=0.099\textheight]{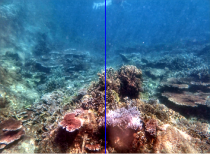}\\
		\includegraphics[width=0.192\textwidth,height=0.093\textheight]{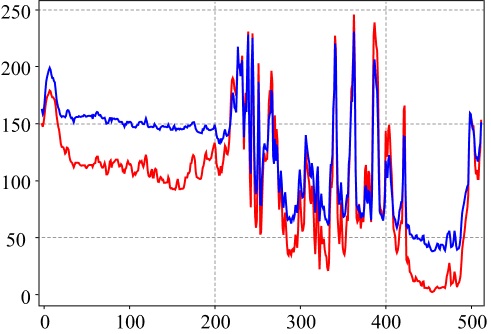}
		&\includegraphics[width=0.192\textwidth,height=0.093\textheight]{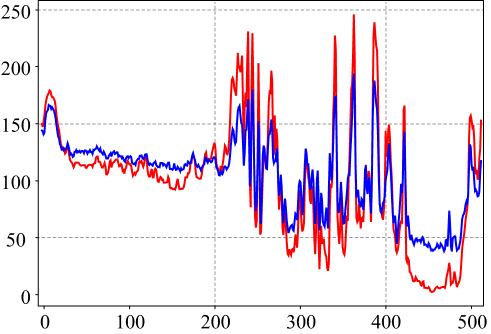}
		&\includegraphics[width=0.192\textwidth,height=0.093\textheight]{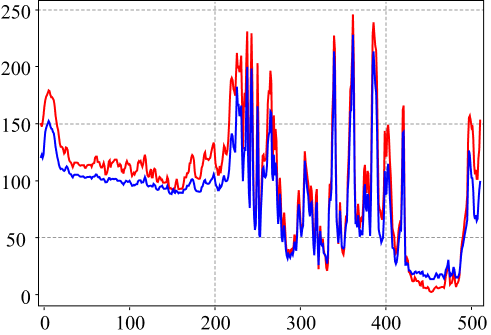}
		&\includegraphics[width=0.192\textwidth,height=0.093\textheight]{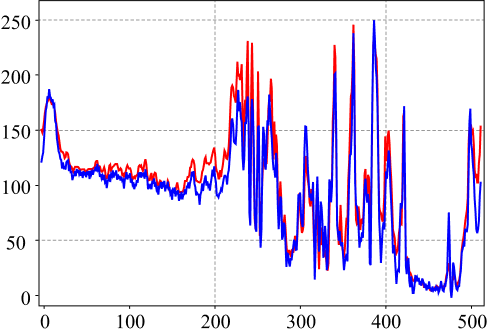}
		&\includegraphics[width=0.192\textwidth,height=0.093\textheight]{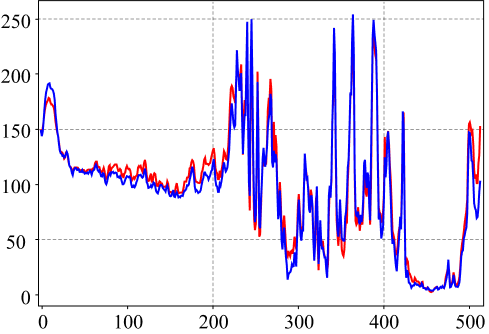}\\
		CLUIE&MBANet&SemiUIR&WaterFlow&Ours
	\end{tabular}
	%	\vspace{-5pt}
	\vspace{5pt}
	\caption{Visual comparison on UIEBD~\citep{li2019waternetuiebd} dataset. We further conduct the pixel distribution of enhanced results and reference images in the uniform region. Obviously, the proposed method performance the best in both visualization and distribution comparison.} 
	\label{fig:UIEBD}
\end{figure*}
\begin{figure*}[!htb]
	\centering
	\setlength{\tabcolsep}{1pt}
	\begin{tabular}{cccccc}
		\includegraphics[width=0.192\textwidth,height=0.097\textheight]{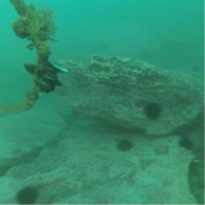}
		&\includegraphics[width=0.192\textwidth,height=0.097\textheight]{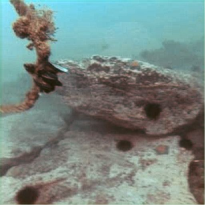}
		&\includegraphics[width=0.192\textwidth,height=0.097\textheight]{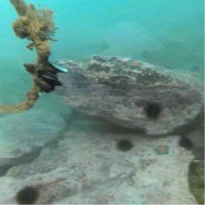}
		&\includegraphics[width=0.192\textwidth,height=0.097\textheight]{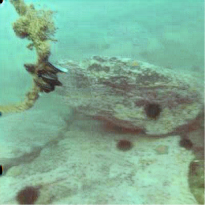}
		&\includegraphics[width=0.192\textwidth,height=0.097\textheight]{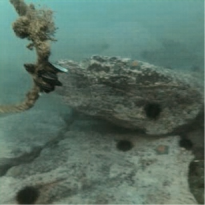}\\
		\includegraphics[width=0.192\textwidth,height=0.095\textheight]{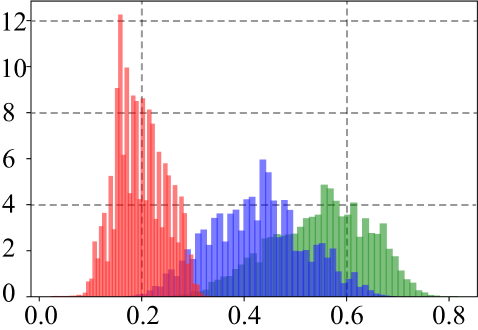}
		&\includegraphics[width=0.192\textwidth,height=0.095\textheight]{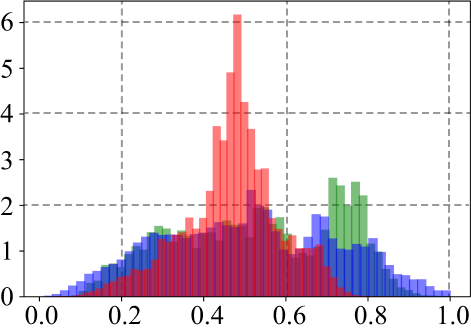}
		&\includegraphics[width=0.192\textwidth,height=0.095\textheight]{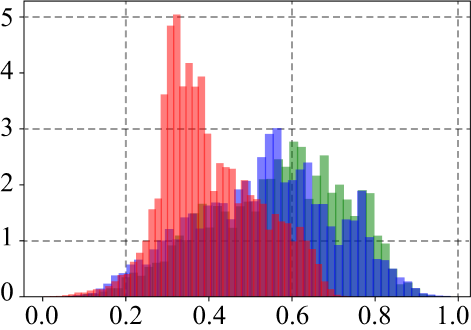}
		&\includegraphics[width=0.192\textwidth,height=0.095\textheight]{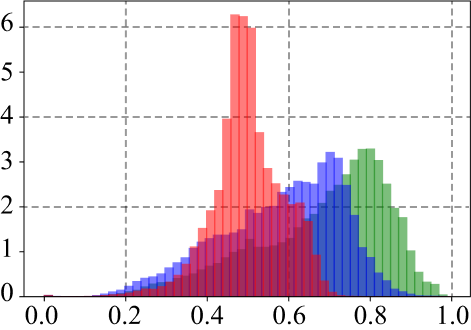}
		&\includegraphics[width=0.192\textwidth,height=0.095\textheight]{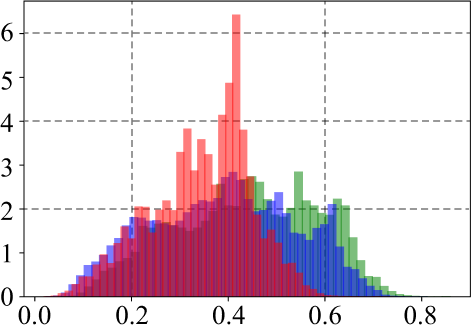}\\
		Input&DLIFM&Ucolor&CWR&TOPAL\\
		\includegraphics[width=0.192\textwidth,height=0.097\textheight]{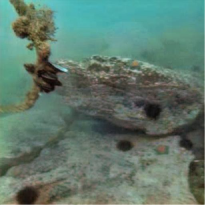}
		&\includegraphics[width=0.192\textwidth,height=0.097\textheight]{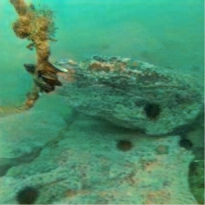}
		&\includegraphics[width=0.192\textwidth,height=0.097\textheight]{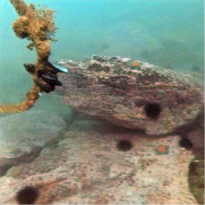}
		&\includegraphics[width=0.192\textwidth,height=0.097\textheight]{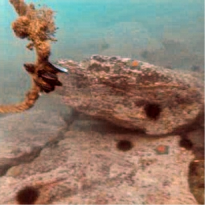}
		&\includegraphics[width=0.192\textwidth,height=0.097\textheight]{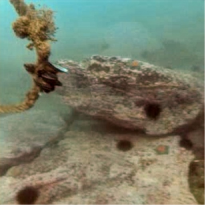}	\\ 
		\includegraphics[width=0.192\textwidth,height=0.095\textheight]{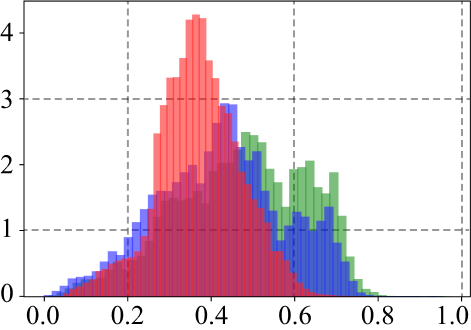}
		&\includegraphics[width=0.192\textwidth,height=0.095\textheight]{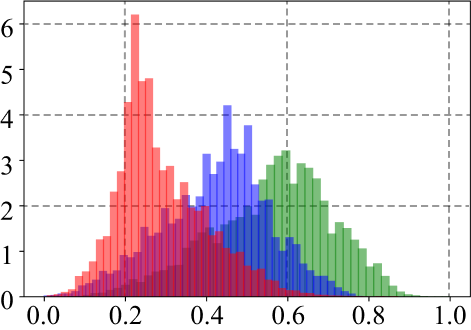}
		&\includegraphics[width=0.192\textwidth,height=0.095\textheight]{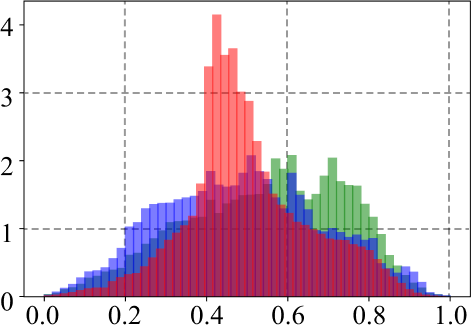}
		&\includegraphics[width=0.192\textwidth,height=0.095\textheight]{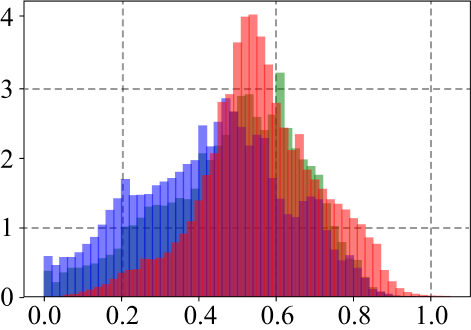}
		&\includegraphics[width=0.192\textwidth,height=0.095\textheight]{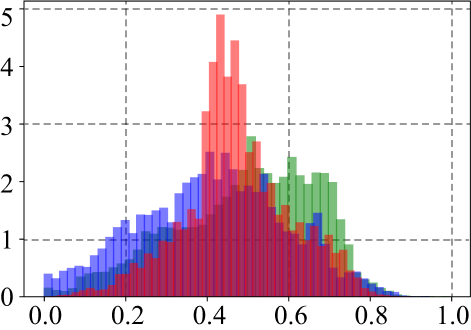}
		%	\\ Input&Water-Net&UWCNN&DLIFM&Ucolor&TOPAL&TACL&Ours\\
		\\CLUIE&MBANet&SemiUIR&WaterFlow&Ours\\
	\end{tabular}
%	\vspace{-0pt}
	\caption{Visualization comparison on UCCS~\citep{liu2020uccs} dataset. We further conduct the histogram distribution of RGB color channel of the dataset. The x and y axis of the histogram respectively represent the pixel intensity and the probability distribution. It is obvious that the image obtained by the proposed method enhances the color distribution closest to the in-air image.} 
	%	The closer the distribution of the three color space, the better the color enhancement effect of the image.}
\label{fig:UCCS}
\end{figure*}
\begin{figure*}[!htb]
\centering
\setlength{\tabcolsep}{1pt}
\begin{tabular}{cccccc}
	\includegraphics[width=0.192\textwidth,height=0.095\textheight]{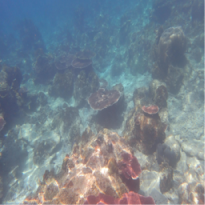}
	&\includegraphics[width=0.192\textwidth,height=0.095\textheight]{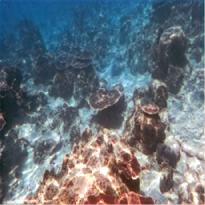}
	&\includegraphics[width=0.192\textwidth,height=0.095\textheight]{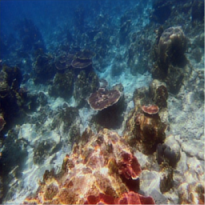}
	&\includegraphics[width=0.192\textwidth,height=0.095\textheight]{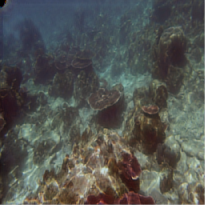}
	&\includegraphics[width=0.192\textwidth,height=0.095\textheight]{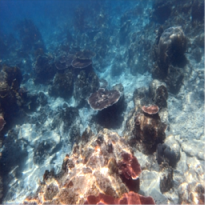}\\
	\includegraphics[width=0.192\textwidth,height=0.092\textheight]{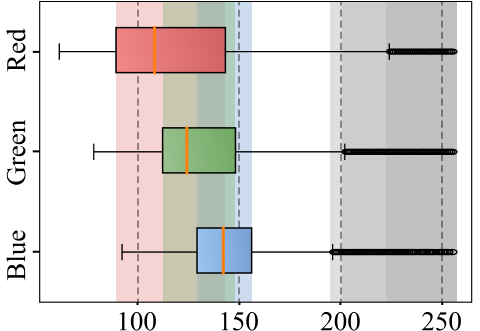}
	&\includegraphics[width=0.192\textwidth,height=0.092\textheight]{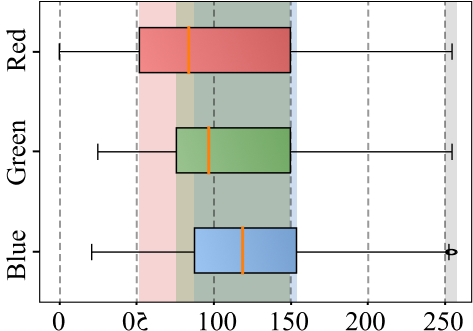}
	&\includegraphics[width=0.192\textwidth,height=0.092\textheight]{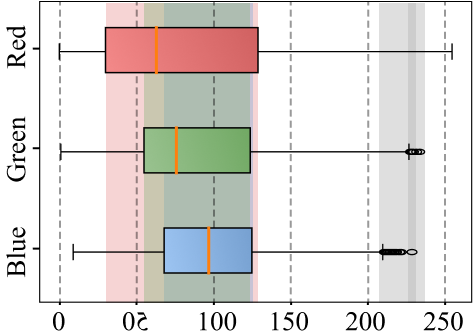}
	&\includegraphics[width=0.192\textwidth,height=0.092\textheight]{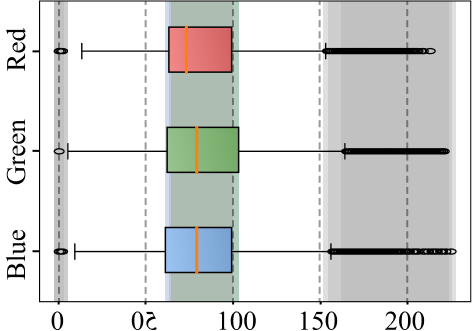}
	&\includegraphics[width=0.192\textwidth,height=0.092\textheight]{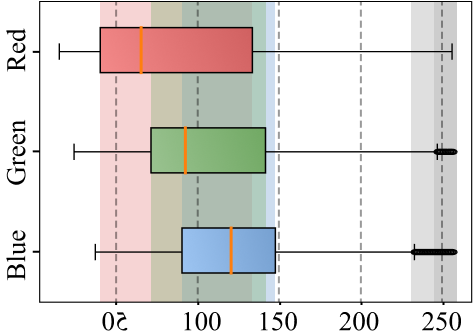}\\
	Input&DLIFM&Ucolor&CWR&TOPAL\\
	\includegraphics[width=0.192\textwidth,height=0.095\textheight]{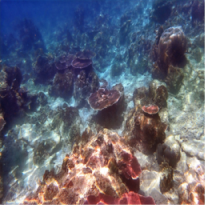}
	&\includegraphics[width=0.192\textwidth,height=0.095\textheight]{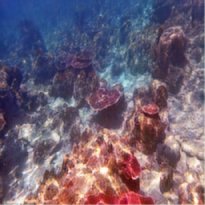}
	&\includegraphics[width=0.192\textwidth,height=0.095\textheight]{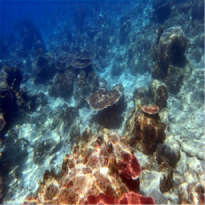}
	&\includegraphics[width=0.192\textwidth,height=0.095\textheight]{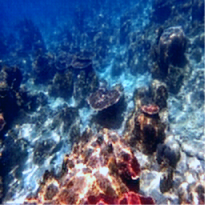}
	&\includegraphics[width=0.192\textwidth,height=0.095\textheight]{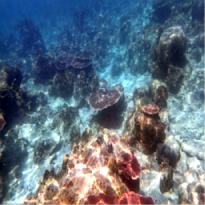}
	\\
	\includegraphics[width=0.192\textwidth,height=0.092\textheight]{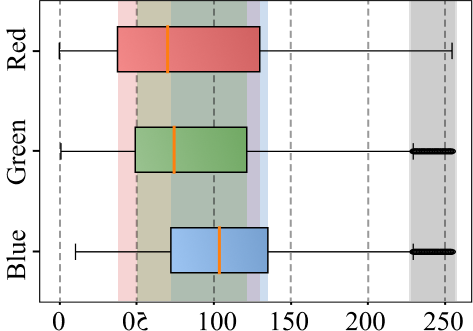}
	&\includegraphics[width=0.192\textwidth,height=0.092\textheight]{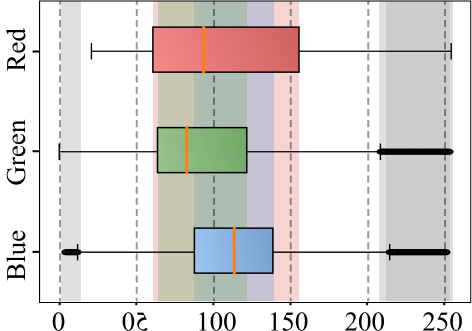}
	&\includegraphics[width=0.192\textwidth,height=0.092\textheight]{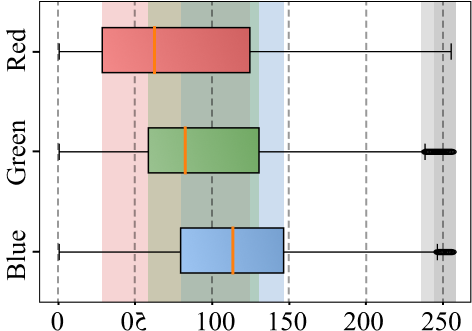}
	&\includegraphics[width=0.192\textwidth,height=0.092\textheight]{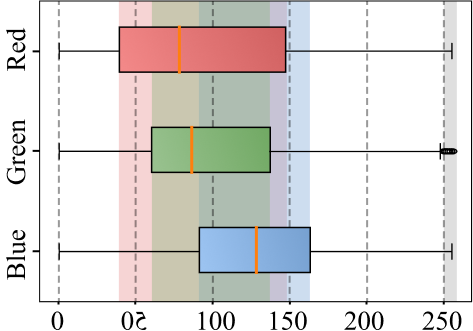}
	&\includegraphics[width=0.192\textwidth,height=0.092\textheight]{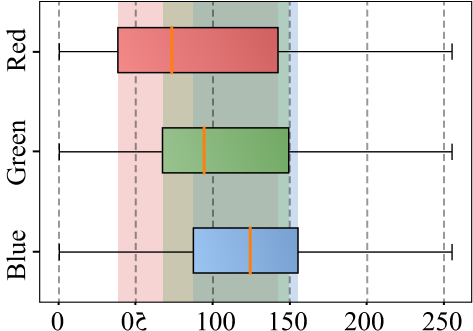}\\
	CLUIE&MBANet&SemiUIR&WaterFlow&Ours\\
\end{tabular}
%\vspace{-2pt}
\caption{Enhancement results on U45~\citep{li2019fusion} dataset. We additionally calculated the dispersion on the RGB color channels of these enhanced images. The x-axis represents the pixel intensity on RGB color channels. Black circles represent outliers that deviate from the mainstream. The appearance of outliers indicates abnormalities in image enhancement. The proposed method achieves robust enhancement while improving the brightness and contrast of the image.}
\label{fig:U45}
\end{figure*}

\begin{figure*}[!htb]
\centering
\setlength{\tabcolsep}{1pt}
\includegraphics[width=0.98\textwidth]{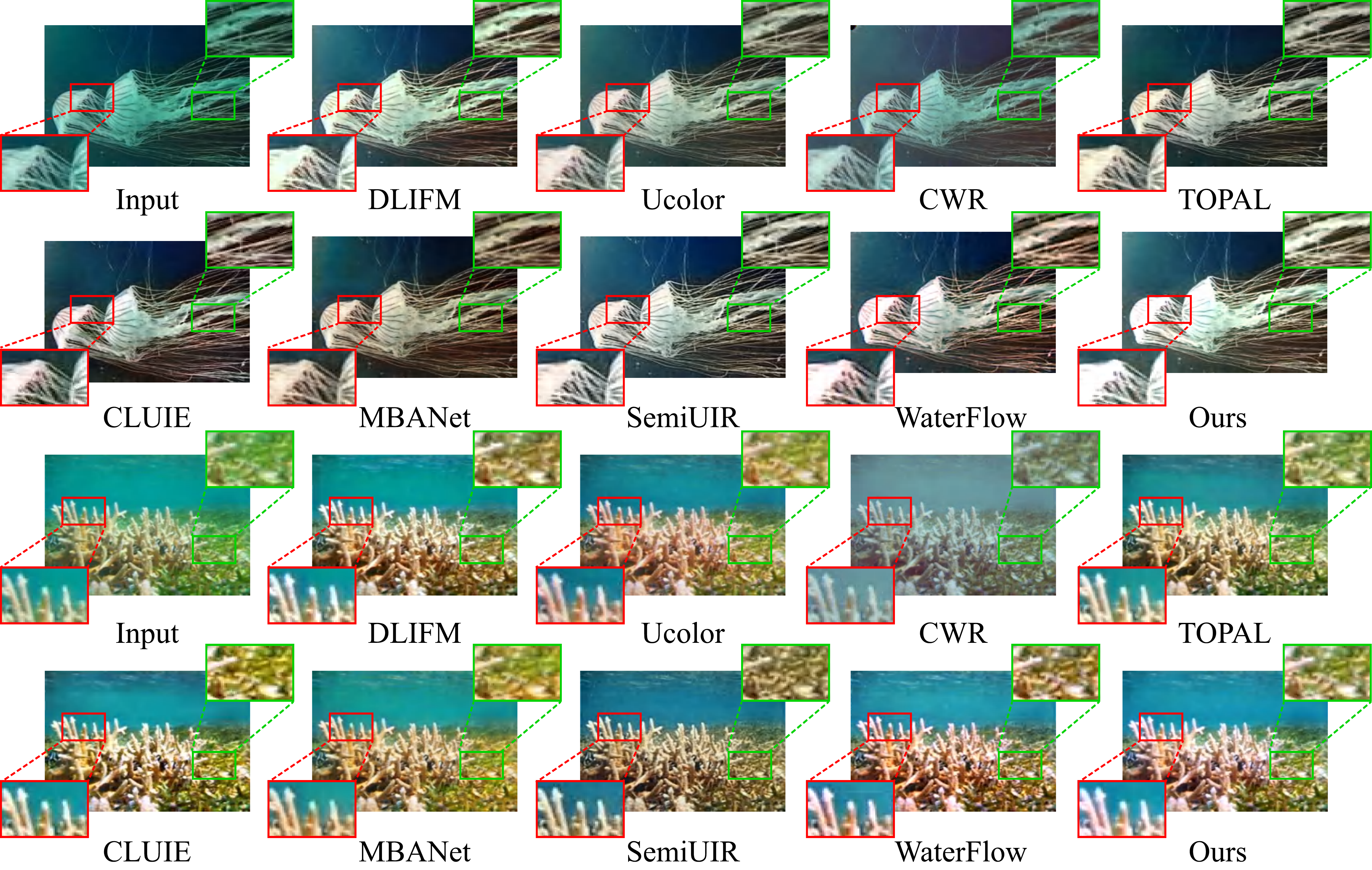}
%\vspace{-8pt}
\caption{Enhancement results on EUVP~\citep{islam2020euvp} dataset. The proposed method outperforms other methods significantly when it comes to rectifying color degradation and improving contrast, particularly in the zoomed-in regions.}
\label{fig:EUVP}
\end{figure*}

\begin{table*}[h]
\renewcommand\arraystretch{1.35}
\caption{Quantitative comparison for underwater image enhancement. $\uparrow$ indicates that higher values correspond to superior outcomes. The top-performing and second-best results are highlighted in \textbf{bold} and \underline{underline}.}
%\vspace{-5pt}
\footnotesize
\setlength{\tabcolsep}{1.05mm}{
	\begin{tabular}{l|ccccc|ccc|ccc|ccc}
		\toprule %???
		\multirow{2}{*}{\textbf{Method}}
		&\multicolumn{5}{c|}{\cellcolor[RGB]{255,204,201}\textbf{UIEBD}}
		&\multicolumn{3}{c|}{\cellcolor[RGB]{154,255,154}\textbf{UCCS}}
		&\multicolumn{3}{c|}{\cellcolor[RGB]{93,173,226}\textbf{U45}}  
		&\multicolumn{3}{c}{\cellcolor[RGB]{225,255,154}\textbf{EUVP}}\\
		&\textbf{PSNR$\uparrow$}&\textbf{SSIM$\uparrow$}&\textbf{UCIQE$\uparrow$}&\textbf{UIQM$\uparrow$}&\textbf{CEIQ$\uparrow$}
		&\textbf{UCIQE$\uparrow$}&\textbf{UIQM$\uparrow$}&\textbf{CEIQ$\uparrow$}
		&\textbf{UCIQE$\uparrow$}&\textbf{UIQM$\uparrow$}&\textbf{CEIQ$\uparrow$}
		&\textbf{UCIQE$\uparrow$}&\textbf{UIQM$\uparrow$}&\textbf{CEIQ$\uparrow$}
		\\
		\midrule %???
		DLIFM    
		& 20.8153            & 0.8755
		& 0.6117             & 4.0016             & 3.3894
		& 0.5196             & 3.5007             & 3.2989
		& 0.5880             & 4.2763             & 3.4242
		& 0.6205             & 4.1051             & 3.4421
		\\Ucolor   
		& 21.0556            & 0.8811
		& 0.5703             & 3.9489             & 3.2727
		& 0.5214             & 3.0376             & 3.2830
		& 0.5641             & \underline{4.3708} & 3.3165
		& 0.5846             & 4.3043             & 3.3318
		\\CWR
		& 17.3749            & 0.7781
		& 0.5149             & 4.1053             & 3.1123
		& 0.4938             & 3.1352             & 3.1866
		& 0.5176             & 4.1906             & 3.1445
		& 0.5212             & 4.1942             & 3.1142
		\\TOPAL  
		& 19.9745            & 0.8787 
		& 0.5744             & 3.9383             & 3.3250  
		& 0.4798             & 3.4701             & 3.1644         
		& 0.5524             & 3.9625             & 3.3525
		& 0.6010             & 4.2689             & 3.3845
		\\CLUIE
		& 19.0104            & 0.8684
		& 0.5853             & 4.0714             & 3.3235      
		& 0.5093             & 3.5045             & 3.1614
		& 0.5721             & 4.1504             & 3.3113
		& 0.6197             & 4.4245             & 3.4273   
		\\MBANet    
		& 18.6140            & 0.8376 
		& 0.5755             & 4.1266             & 3.3861
		& 0.5043             & 2.9082             & 3.2040
		& 0.5602             & 4.2722             & 3.2400
		& 0.6061             & \underline{4.5741} & 3.3585   
		\\SemiUIR    
		& \textbf{22.6627}   & \underline{0.9039} 
		& \underline{0.6188} & 4.1399             & \underline{3.4221}
		& \textbf{0.5519}    & \underline{4.1333} &\textbf{3.3663}
		& 0.5880             & 4.2766             & 3.4850
		& 0.6183             & 4.4189             & 3.4600
		\\WaterFlow    
		& 22.4815            & 0.8786
		& 0.6046             & \underline{4.1885} & 3.4099
		& 0.5354             & 4.0905             & 3.3214
		& \underline{0.6229} & \textbf{4.4129}    & \underline{3.5003}
		& \textbf{0.6397}    & 4.4400             & \underline{3.4679} 
		\\Ours     
		& \underline{22.5480}& \textbf{0.9056} 
		& \textbf{0.6196}    & \textbf{4.2008}    & \textbf{3.4291}
		& \underline{0.5381} & \textbf{4.6994}    & \underline{3.3278}
		& \textbf{0.6241}    & 4.2240             & \textbf{3.5180}
		& \underline{0.6300} & \textbf{20.9151}   & \textbf{3.4769}\\
		\bottomrule %???
\end{tabular}}
\label{tab:evaluation_enhance}
\end{table*}

\section{Experiments}

\subsection{Datasets and Metrics}
In this section, we evaluate the effectiveness of the proposed method through qualitative and quantitative comparison. Specifically, widely used UIEBD~\citep{li2019waternetuiebd},  UCCS~\citep{liu2020uccs}, U45~\citep{li2019fusion} and EUVP~\citep{islam2020euvp} datasets are used to evaluate the performance of HUPE for underwater image enhancement. 
UIEBD includes 890 images with the corresponding reference images chosen with laborious and well-designed pairwise comparisons from 12 enhancement methods. They are divided into 90 image pairs for testing and 800 image pairs for training. 
The UCCS dataset was used to evaluate the enhancement ability of correcting color casts, which contain 300 images in blue, green and  blue-green tones.
U45 includes 45 underwater images, which contains underwater degraded color casts, low contrast, and haze effects. The EUVP dataset contains 675 pairs of underwater images, which are captured by 7 different cameras or carefully selected from the internet to adapt to the natural variations of different underwater scenes. It is worth mentioning, considering that the calculation of FFT only supports powers of 2 signal length. In subsequent comparisons, we will uniformly resize UIEBD and UCCS datasets to 512$\times$512.

\subsection{Implement Details}
Our network is implemented using PyTorch and trained on an NVIDIA RTX 3090 GPU. Specifically, we first separately train the Heuristic Invertible Network and the Task Perception Module. During the training of the Heuristic Invertible Network, we set the number of epochs to 100, use the Adam optimizer with a learning rate of 1e-5 and a batch size of 1. UIEBD~\citep{li2019waternetuiebd} dataset is utilized for training, which consists of 800 images. During training, we randomly crop images of size $512 \times 512$ from the training data for augmentation. When training the task-aware module, SSD~\citep{liu2016ssd}, Refinedet~\citep{zhang2018refinedet}, PAA~\citep{kim2020paa} serve as the backbones of the Task Perception Module for object detection, while LEDNet~\citep{wang2019lednet} and CGNet~\citep{wu2020cgnet} serve as the backbones for semantic segmentation. The RUIE~\citep{liu2020uccs} dataset and Aquarium~\citep{aquarium} dataset are used as underwater object detection datasets for training. The initial learning rate for the detection task is set to 1e-2, with a decay weight of 1e-4, a batch size of 1, and the optimizer used is SGD. SUIM~\citep{2020SUIM} is used as the training dataset for underwater semantic segmentation, with an initial learning rate of 1e-3, a decay factor of 5e-4, and a batch size of 7. After training these networks separately, we jointly finetune the pre-trained Heuristic Invertible Network and the Task Perception Module for 20 epochs. The values of $\lambda_1$, $\lambda_2$, $\lambda_3$, and $\lambda_4$ are set to $1$, $0.05$, $1$, and $0.2$.

The RUIE~\citep{liu2020uccs} and Aquarium~\citep{aquarium} datasets for underwater object detection and SUIM~\citep{2020SUIM} for underwater semantic segmentation were used to evaluate the effectiveness of Tatarflow in subsequent perception tasks. RUIE contains 2070 training images and 676 testing images captured from underwater scenes with abundant underwater lives. The corresponding detection labels consist of urchin, trepang and scallop. Aquarium contains 448 training images and 63 testing images captured in the aquarium with multiple underwater species, including fish, penguins, jellyfish, puffins, sharks, stingrays and starfish.
SUIM contains 1525 training images and 110 testing images for semantic segmentation including aquatic plants and sea-grass~(PF), Robots~(RO), wrecks and ruins~(WR), human divers~(HD), sea-floor and rocks~(SR), reefs and invertebrates~(RI) and background~(BW).

Considering the lack of ground-truth in most underwater image datasets, we further employ non-reference evaluation metrics such as Underwater Color Image Quality Evaluation~(UCIQE)~\citep{yang2015uciqe},  Underwater Image Quality Measurement~(UIQM)~\citep{panetta2015uiqmuism}, and Contrast Enhancement based Contrast-changed Image Quality Measure~(CEIQ)~\citep{fang2014ceiq} to assess the effectiveness of the proposed method in the underwater image enhancement task. 
UCIQE is a metric that measures color cast, blurring, and contrast deficiencies in captured underwater images, offering a simple and fast solution for real-time image processing with improved correlation to subjective vision perception evaluation. 
UIQM is proposed to comprehensively evaluate whether the image conforms to human perception through sharpness, contrast, and color. CEIQ considers contrast distortion in underwater images based on the principle of natural scene statistics. Higher scores of these metrics represent the better visual quality of the enhancement results.

UIEBD provides reference images that are obtained and selected from different underwater image enhancement methods. We additionally use the fully-reference metrics Peak Signal to Noise Ratio (PSNR) and Structural Similarity (SSIM)~\citep{wang2004image} on UIEBD to evaluate the similarity between the enhancement image and the reference image. PSNR utilizes the maximum and mean square error to express the peak signal-to-average energy ratio. SSIM assesses brightness, contrast, and structure to quantify the similarity between the results and reference images. Higher scores of the above metrics indicate a higher degree of similarity between the output and the reference image in both structure and content. The commonly utilized Average Precision (AP) and Intersection over Union (IoU) are adopted as the task-driven evaluation metric in detection and segmentation tasks, which are both positive for the performance of subsequent perception tasks.

\begin{figure*}[!htb]
	\centering
	\setlength{\tabcolsep}{1pt}
	\begin{tabular}{ccccc}
		\includegraphics[width=0.192\textwidth]{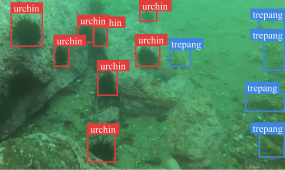}
		&\includegraphics[width=0.192\textwidth]{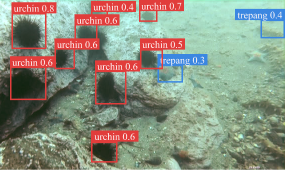}
		&\includegraphics[width=0.192\textwidth]{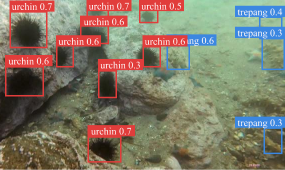}
		&\includegraphics[width=0.192\textwidth]{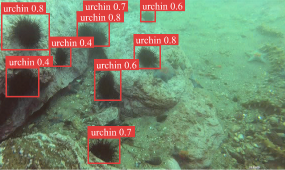}
		&\includegraphics[width=0.192\textwidth]{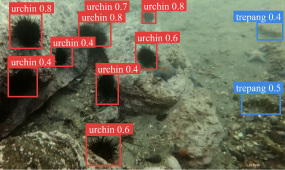}
		\\Ground Truth&DLIFM&Ucolor&CWR&TOPAL\\
		\includegraphics[width=0.192\textwidth]{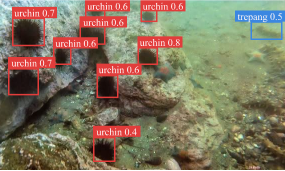}
		&\includegraphics[width=0.192\textwidth]{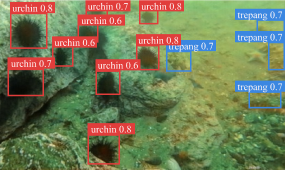}
		&\includegraphics[width=0.192\textwidth]{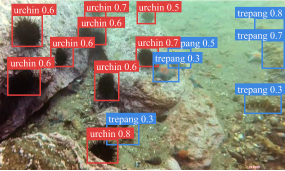}
		&\includegraphics[width=0.192\textwidth]{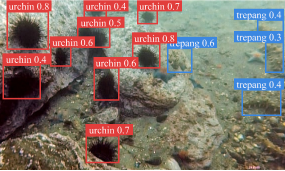}
		&\includegraphics[width=0.192\textwidth]{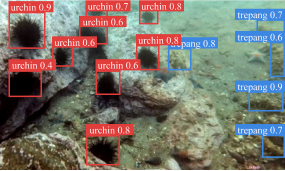}
		\\CLUIE&MBANet&SemiUIR&WaterFlow& Ours\\
	\end{tabular}
%	\vspace{-5pt}
	\caption{Assessment of detection performance on RUIE~\citep{liu2020uccs} dataset. It is obvious that the proposed HUPE performs best in both accuracy and quantity.}
	\label{fig:UCCSdetection}
\end{figure*}

\begin{figure*}[h]
	\centering
	\begin{minipage}{0.35\textwidth}
		
		\subfloat{
			\begin{minipage}{1\textwidth}
				\includegraphics[width=1\textwidth,height=0.243\textheight]{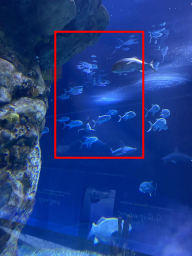}
				\centering   Input
			\end{minipage}
		}
	\end{minipage}
	\begin{minipage}{0.120\textwidth}
		\subfloat{
			\begin{minipage}{1\textwidth}
				\vspace{-1pt}
				\includegraphics[width=1\textwidth,height=0.108\textheight]{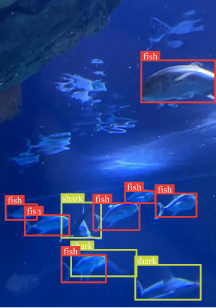}
				\centering    Ground truth\vspace{-6pt}  	
			\end{minipage}
		}\quad    
		\subfloat{
			
			\begin{minipage}{1\textwidth}
				\includegraphics[width=1\textwidth,height=0.107\textheight]{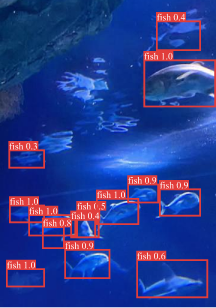}
				\centering  CLUIE
			\end{minipage}
		}
	\end{minipage}
	\begin{minipage}{0.120\textwidth}
		\subfloat{
			\begin{minipage}{1\textwidth}\vspace{-1pt}
				\includegraphics[width=1\textwidth,height=0.108\textheight]{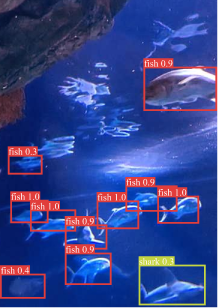}
				\centering  DLIFM\vspace{-6pt}  	
			\end{minipage}
		}\quad  
		\subfloat{
			\begin{minipage}{1\textwidth}
				\includegraphics[width=1\textwidth,height=0.107\textheight]{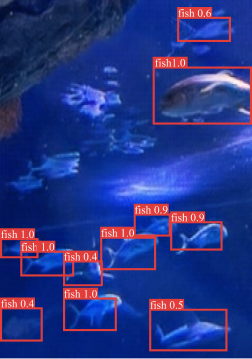}
				\centering  MBANet
			\end{minipage}
		}
	\end{minipage}
	\begin{minipage}{0.120\textwidth}
		\subfloat{
			\begin{minipage}{1\textwidth}\vspace{-1pt}
				\includegraphics[width=1\textwidth,height=0.108\textheight]{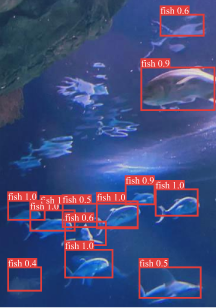}
				\centering  Ucolor\vspace{-6pt}  	
			\end{minipage}
		}\quad  
		\subfloat{
			\begin{minipage}{1\textwidth}
				\includegraphics[width=1\textwidth,height=0.107\textheight]{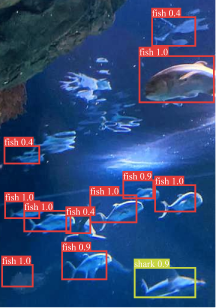}
				\centering  SemiUIR
			\end{minipage}
		}
	\end{minipage}
	\begin{minipage}{0.120\textwidth}
		\subfloat{
			\begin{minipage}{1\textwidth}\vspace{-1pt}
				\includegraphics[width=1\textwidth,height=0.108\textheight]{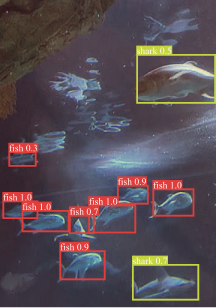}
				\centering  CWR \vspace{-6pt}  	
			\end{minipage}
		}\quad  
		\subfloat{
			\begin{minipage}{1\textwidth}
				\includegraphics[width=1\textwidth,height=0.107\textheight]{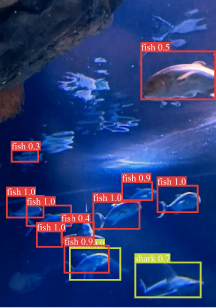}
				\centering  WaterFlow
			\end{minipage}
		}
	\end{minipage}
	\begin{minipage}{0.120\textwidth}\vspace{-1pt}
		\subfloat{
			\begin{minipage}{1\textwidth}
				\includegraphics[width=1\textwidth,height=0.108\textheight]{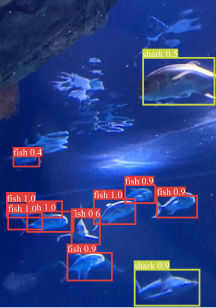}
				\centering  TOPAL \vspace{-6pt}  	
			\end{minipage}
		}\quad  
		\subfloat{
			\begin{minipage}{1\textwidth}
				\includegraphics[width=1\textwidth,height=0.107\textheight]{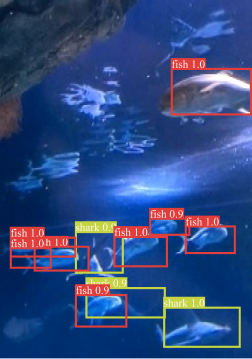}
				\centering  Ours
			\end{minipage}
		}
	\end{minipage}
	\caption{Evaluation of object detection on Aquarium~\citep{aquarium} dataset with different enhancement methods. The proposed method evidently most suitable for detection tasks.}
	\label{fig:aquarium}
\end{figure*}

\begin{figure*}[h]
	\centering
	\begin{minipage}{0.30\textwidth}
		
		\subfloat{
			\begin{minipage}{1\textwidth}
				\includegraphics[width=1\textwidth,height=0.177\textheight]{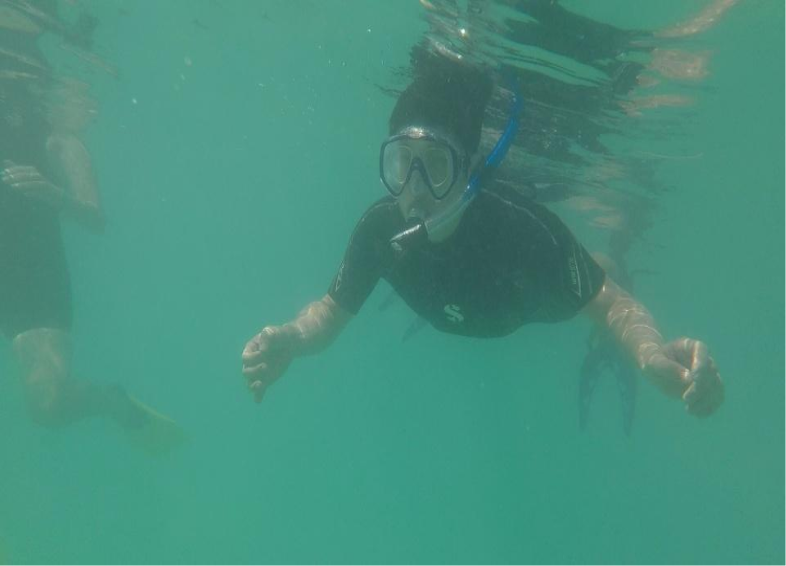}
				\centering   Input
			\end{minipage}
		}
	\end{minipage}
	\begin{minipage}{0.130\textwidth}
		\subfloat{
			\begin{minipage}{1\textwidth}
				\vspace{-1pt}
				\includegraphics[width=1\textwidth,height=0.076\textheight]{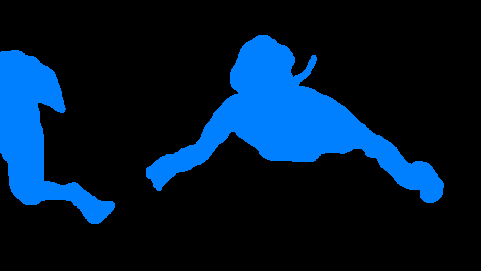}
				\centering    Ground truth\vspace{-6pt}  	
			\end{minipage}
		}\quad    
		\subfloat{
			
			\begin{minipage}{1\textwidth}
				\includegraphics[width=1\textwidth,height=0.074\textheight]{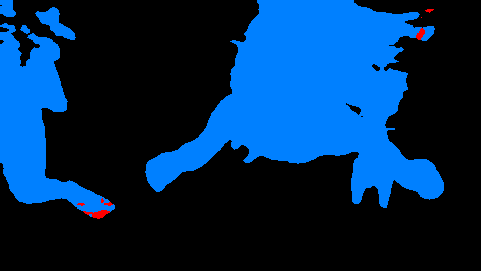}
				\centering  CLUIE
			\end{minipage}
		}
	\end{minipage}
	\begin{minipage}{0.130\textwidth}
		\subfloat{
			\begin{minipage}{1\textwidth}\vspace{-1pt}
				\includegraphics[width=1\textwidth,height=0.076\textheight]{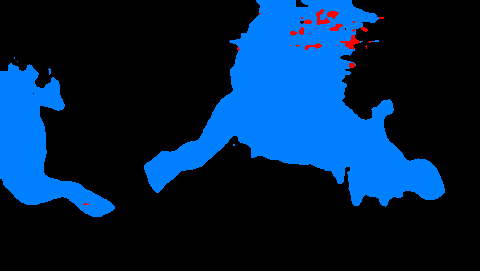}
				\centering  DLIFM\vspace{-6pt}  	
			\end{minipage}
		}\quad  
		\subfloat{
			\begin{minipage}{1\textwidth}
				\includegraphics[width=1\textwidth,height=0.074\textheight]{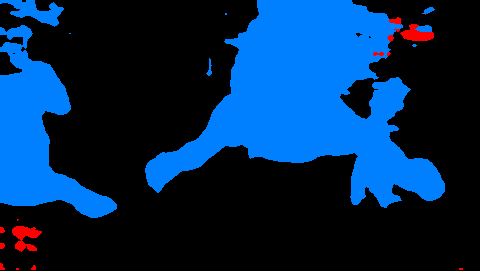}
				\centering  MBANet
			\end{minipage}
		}
	\end{minipage}
	\begin{minipage}{0.130\textwidth}
		\subfloat{
			\begin{minipage}{1\textwidth}\vspace{-1pt}
				\includegraphics[width=1\textwidth,height=0.076\textheight]{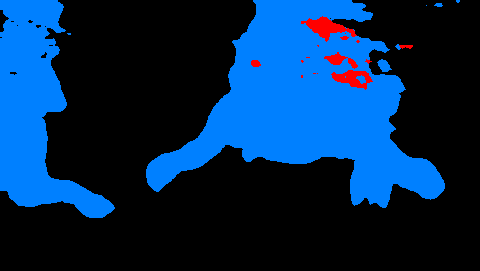}
				\centering  Ucolor\vspace{-6pt}  	
			\end{minipage}
		}\quad  
		\subfloat{
			\begin{minipage}{1\textwidth}
				\includegraphics[width=1\textwidth,height=0.074\textheight]{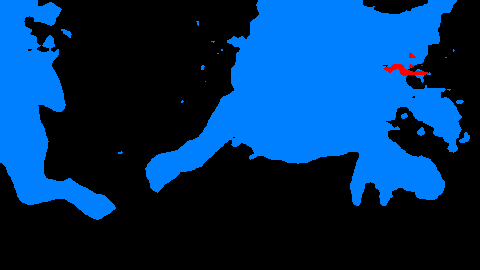}
				\centering  SemiUIR
			\end{minipage}
		}
	\end{minipage}
	\begin{minipage}{0.130\textwidth}
		\subfloat{
			\begin{minipage}{1\textwidth}\vspace{-1pt}
				\includegraphics[width=1\textwidth,height=0.076\textheight]{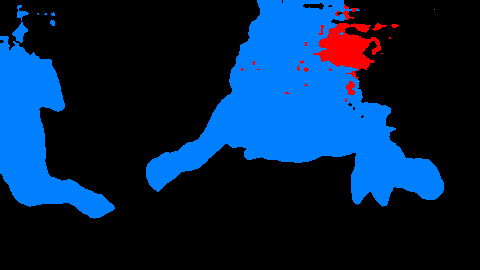}
				\centering  CWR \vspace{-6pt}  	
			\end{minipage}
		}\quad  
		\subfloat{
			\begin{minipage}{1\textwidth}
				\includegraphics[width=1\textwidth,height=0.074\textheight]{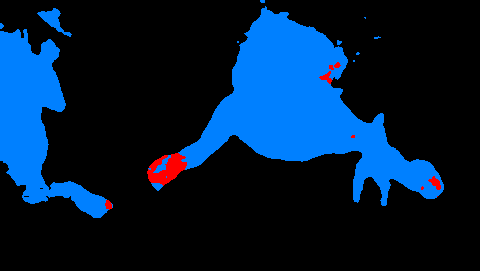}
				\centering  WaterFlow
			\end{minipage}
		}
	\end{minipage}
	\begin{minipage}{0.130\textwidth}\vspace{-1pt}
		\subfloat{
			\begin{minipage}{1\textwidth}
				\includegraphics[width=1\textwidth,height=0.076\textheight]{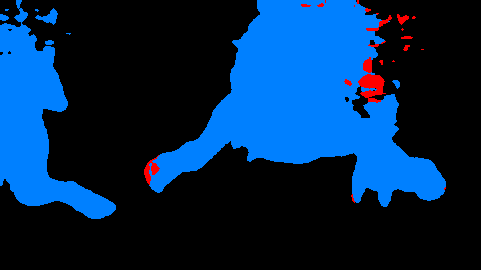}
				\centering  TOPAL \vspace{-6pt}  	
			\end{minipage}
		}\quad  
		\subfloat{
			\begin{minipage}{1\textwidth}
				\includegraphics[width=1\textwidth,height=0.074\textheight]{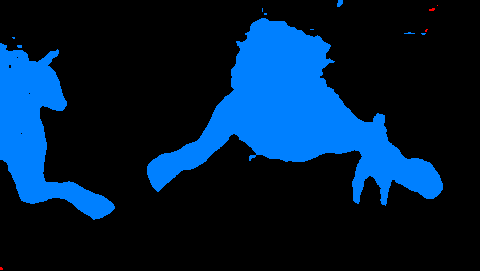}
				\centering  Ours
			\end{minipage}
		}
	\end{minipage}
	\caption{Evaluation of semantic segmantation on SUIM~\citep{2020SUIM} dataset with representative enhancement methods. The proposed method evidently performs best in semantic segmantation tasks.}
	\label{fig:suim}
\end{figure*}

\begin{figure*}[!htb]
	\centering
	\setlength{\tabcolsep}{1pt}
	\includegraphics[width=0.990\textwidth]{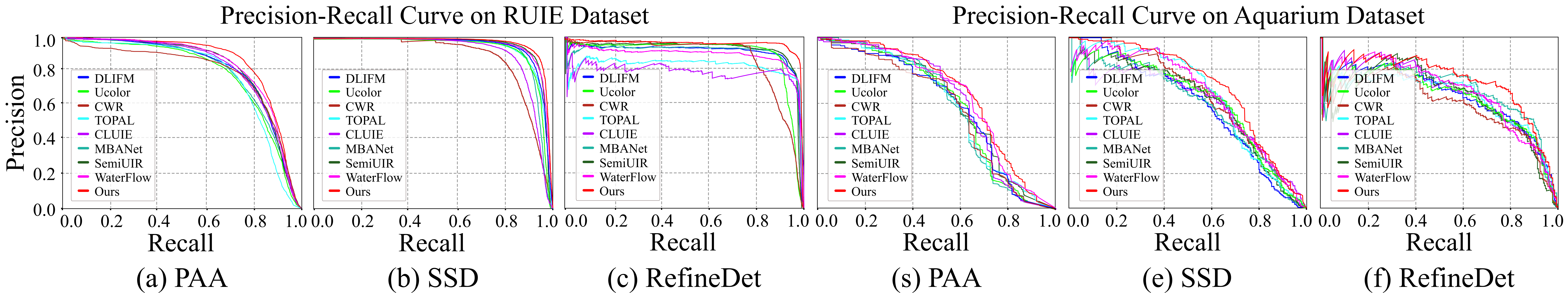}
%	\vspace{-5pt}
	\caption{Precision-Recall and ROC Curve on RUIE~\citep{liu2020uccs}(a)(b) and Aquarium~\citep{aquarium}(c)(d) datasets with the enhanced images generating from the different enhancement methods.}
	%	The closer the distribution of the three color space, the better the color enhancement effect of the image.}
\label{fig:detetion_duibi2}
\end{figure*}

\subsection{Evaluation on Underwater Image Enhancement}
\subsubsection{Qualitative Results}

In this section, we employ experiments on various datasets to assess the performance of the proposed HUPE with eight representative enhancement methods for comparison, including DLIFM~\citep{chen2021dlifm}, Ucolor~\citep{li2021ucolor}, CWR~\citep{han2022CWR}, TOPAL~\citep{jiang2022topal}, CLUIE~\citep{li2022cluie}, MBANet~\citep{xue2023MBANet}, SemiUIR~\citep{huang2023contrastive} and Waterflow~\citep{zhang2023waterflow}. Both visual effects and objective evaluations are analyzed in this section.

For visual comparison, Fig.~\ref{fig:UIEBD} shows the enhancement results of all methods on UIEBD dataset. 
It is evident that the absorption and scattering of light underwater lead to the appearance of undesirable blue hues. 
Ucolor, CWR and TOPAL still exhibit significant color distortion. 
CLUIE and MBANet additionally introduce artifacts and unnatural colors. 
DLIFM, SemiUIR and WaterFlow recover image contrast and brightness but retain noticeable color shifts especially on nearby corals. Only our proposed method effectively mitigates the adverse color bias while restoring bright reflections. 
We additionally provide line charts representing the similarity between the images and reference images. The blue vertical line in the image denotes the calculation range of the line charts. The blue line in the chart represents the pixel intensity of the current image, while the red line represents the pixel intensity of the reference image. It is evident that our results are visually and objectively the closest to the reference image.

Fig.~\ref{fig:UCCS} illustrates the enhancement results on the UCCS dataset. Unlike Fig.~\ref{fig:UIEBD}, the decay of green light is significantly weaker than that of red and blue light in this environment.
Ucolor, TOPAL and CLUIE fail to alleviate the scattering of light underwater. 
CWR and MBANet introduce color artifacts deviating from the original image colors, tending towards a greenish bias. 
SemiUIR and WaterFlow exhibit excessive red color compensation on the rock.
%Although DLIFM produces pleasing images, the instructive details of the image have not been recovered well. 
Compared with the other methods, the proposed method not only effectively corrects the color deviation, but also restores the scene radiance and contrast of the image. 
Fig.~\ref{fig:UCCS} also shows the distribution of RGB color space for the images obtained by different methods. In general, the proposed method more effectively addresses the swift attenuation of red wavelengths underwater, with its color space distribution closely approximating that of in-air images.

Fig.~\ref{fig:U45} depicts the performance of the U45 dataset. In this scenario, underwater imaging is hindered by the presence of fine underwater particles that obscure the color and clarity of the input images, resulting in some color bias. 
%MIP and TOPAL neglect the restoration of underwater image clarity. 
Ucolor, CWR and TOPAL make erroneous estimations of the underwater scene, leading to darker output results. 
MBANet, SemiUIR and WaterFlow overlook the attenuation of different wavelength light in water, resulting in red, green and blue color casts respectively.
%DLIFM introduces excessive red compensation. 
Compared with DLIFM and CLUIE, which also effectively restore image clarity and brightness, the proposed method achieves a relatively better color appearance.
Additionally, we provide box plots representing the pixel intensity distribution of the images below each image, where the x-axis represents pixel intensity. Transparent rectangles have been added to visually illustrate the distribution of intensity in the three color channels and the presence of outliers. 
%It is evident that MIP, DLIFM, and TOPAL exhibit relative neglect of the recovery of the underwater blue channel. 
All comparable  methods exhibit numerous outliers during the restoration process. In contrast, our method not only avoids the generation of outliers but also restores brighter scenes.
\begin{figure*}[!htb]
	\centering
	\setlength{\tabcolsep}{1.0pt}
	\includegraphics[width=1.0\textwidth]{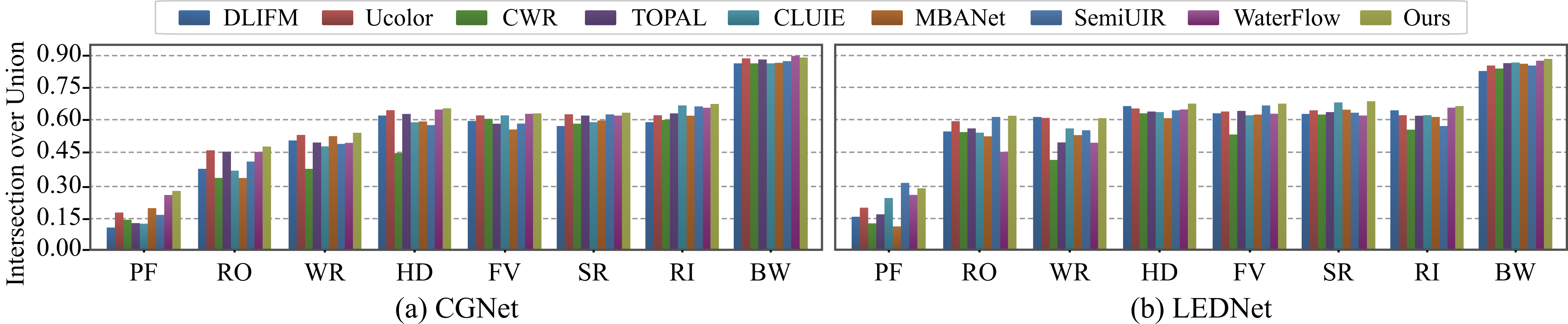}
%	\vspace{-17pt}
	\caption{Evaluation of Intersection ove Union(IoU) on SUIM~\citep{2020SUIM} datasets with the enhanced images generating from the representative methods. The x-axis represents the category of the label.}
	\label{fig:segmentation_duibi2}
\end{figure*}

\begin{figure*}[!htb]
	\centering
	\setlength{\tabcolsep}{1pt}
	\includegraphics[width=0.99\textwidth]{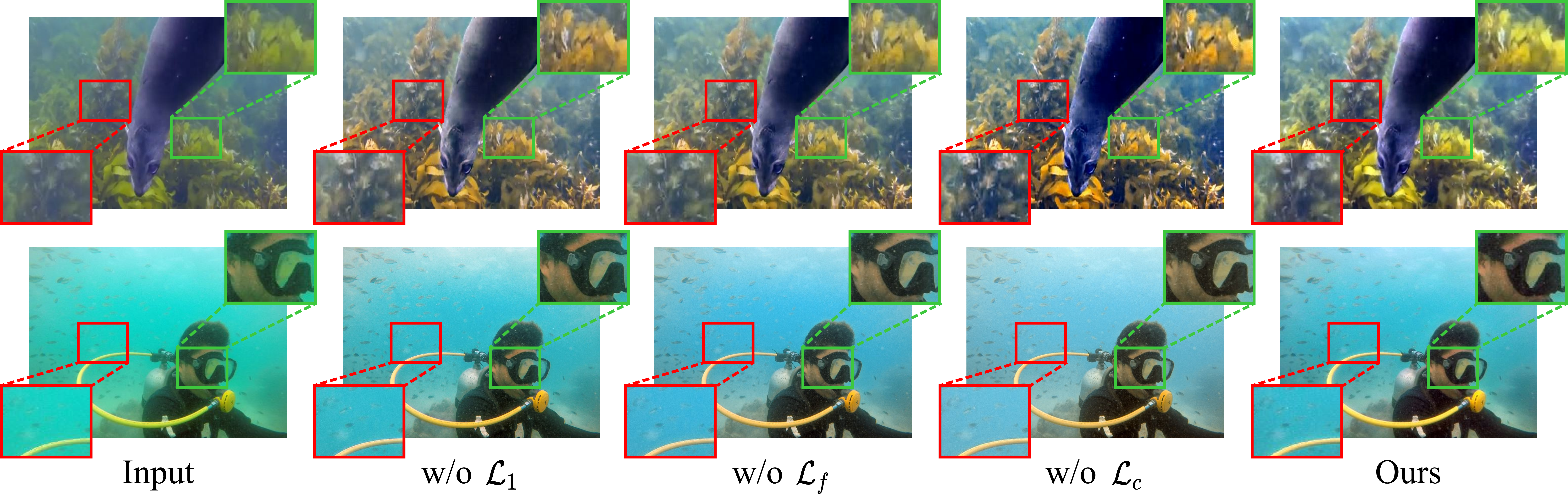}
%	\vspace{-5pt}
	\caption{Ablation on the loss function.~$\mathcal{L}_\mathrm{1}$, $\mathcal{L}_\mathrm{s}$ and  $\mathcal{L}_\mathrm{c}$ respectively denote L1 loss, style loss, and contrasive loss.}
	\label{fig:visualizationforlossablation}
\end{figure*}

\begin{figure*}[!htb]
	\centering
	\setlength{\tabcolsep}{0.10pt}
	\includegraphics[width=0.9\textwidth]{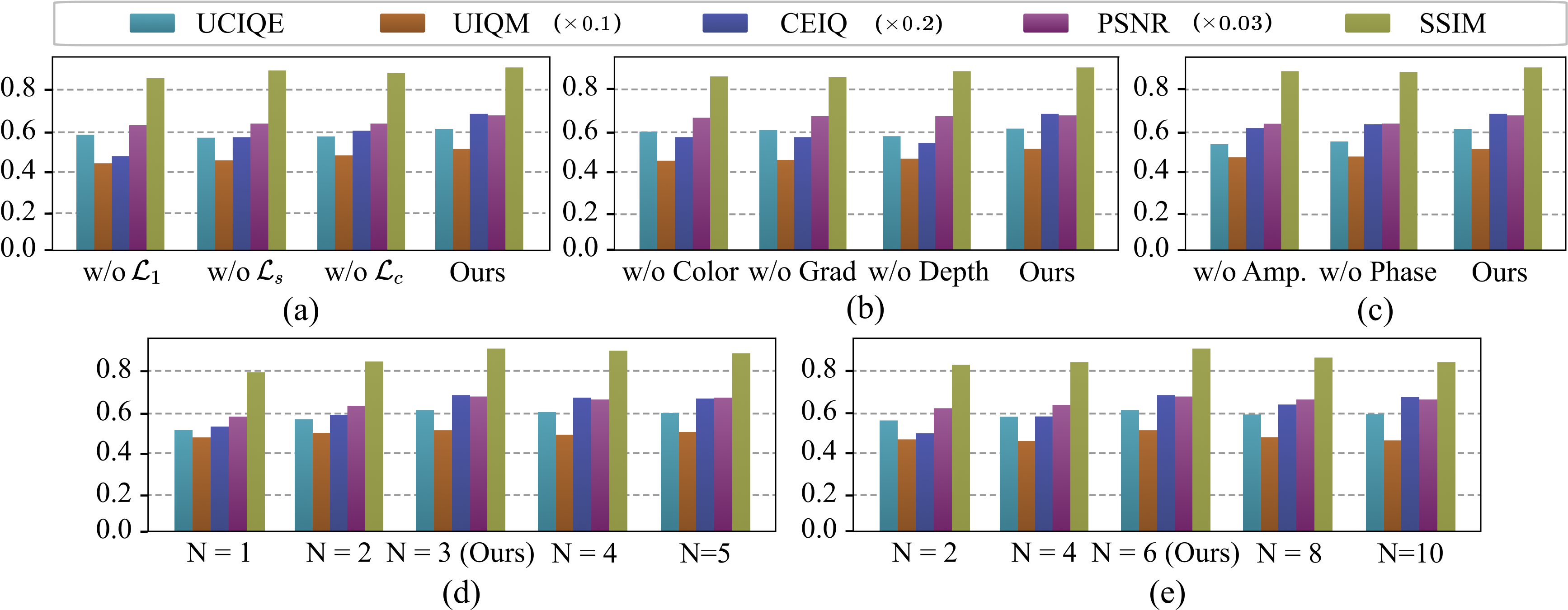}
	%	\vspace{-7pt}
	\caption{Quantitative results of the ablation experiment. (a) denotes the study on the loss function. (b) represents the study for the different input of HPE. (c) represents the study for SFA. (d) represents the study for number of Hybrid Invertible Blocks. (e) represents the study for number of flow steps in each HIB.}
	\label{fig:lossablation}
\end{figure*}
\begin{figure*}[!htb]
	\centering
	\setlength{\tabcolsep}{1pt}
	\includegraphics[width=0.99\textwidth]{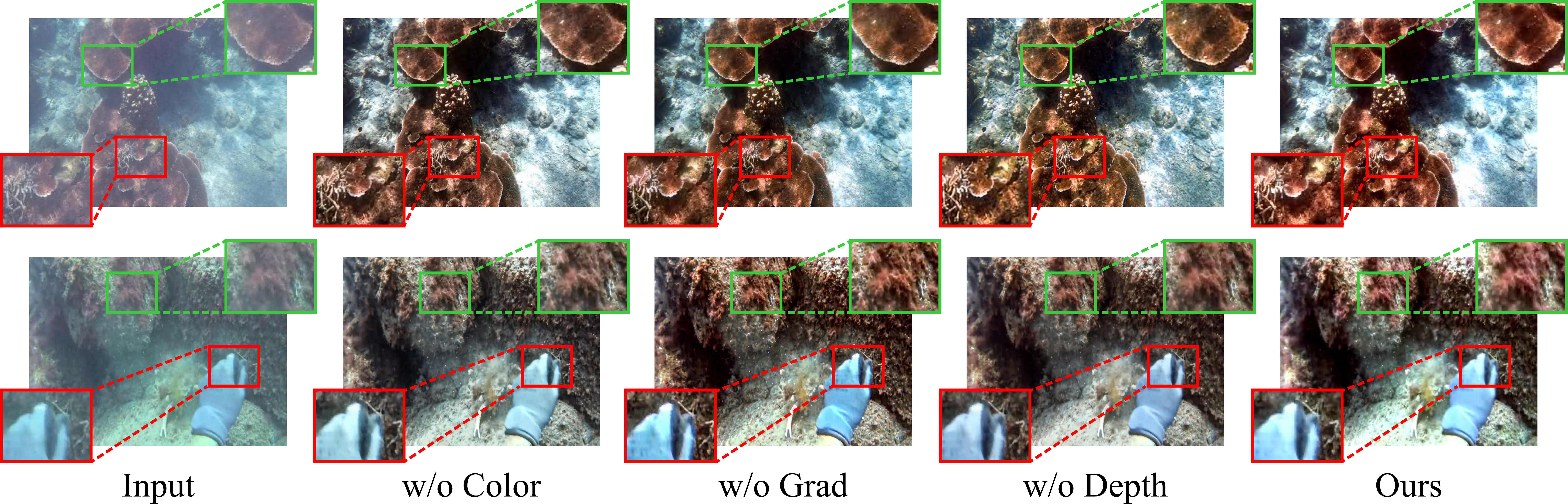}
	%	\vspace{-2pt}
	\caption{Visualization for ablation study on the input of Heuristic Prior guided Encoder.}
	\label{fig:visualizationforinputencoder}
\end{figure*}

Fig.~\ref{fig:EUVP} illustrates the visual comparison of the EUVP dataset. 
It can be clearly seen that white jellyfish endure obvious blue color cast underwater in the first example. CWR exhibits weak color recovery. The jellyfish in DLIFM and Ucolor still suffer from obvious blue color cast. CLUIE, SemiUIR and WaterFlow demonstrate excessive red channel enhancement, especially on the head of the jellyfish. Compared with other methods, our method retains the original color of the jellyfish and also effiectively restores the brightness and contrast. 
In the second exapmle, CWR, DLIFM, SemiUIR, and WaterFlow misestimated the scene reflection, causing the water plants in the green frame to deviate from their original colors. Ucolor, TOPAL, CLUIE and MBANet achieve close-to-natural color enhancement effects, but the restoration of the contrast and sharpness is limited compared to the proposed method.
\subsubsection{Quantitative Results}
Tab.~\ref{tab:evaluation_enhance} presents quantitative results for representative methods of underwater image enhancement. We first provided a supervised evaluation of our method with respect to the reference images provided by UIEBD. Our method achieved the highest SSIM score, but fell below SemiUIR in terms of PSNR. However, it's important to note that the reference images from UIEBD were obtained from various underwater image enhancement methods, not from real-world captures~\citep{li2019waternetuiebd}. Therefore, considering only supervised metrics may not provide a comprehensive assessment of the robustness of underwater image enhancement. Furthermore, we also compared image quality metrics across the datasets mentioned above.
When combining non-reference image quality assessment and full-reference benchmarks, only our proposed method demonstrated outstanding enhancement performance and high robustness, consistent with qualitative results.

\subsection{Evaluation on Subsequent Perception Tasks}
We compared different enhancement methods with representative detection methods and segmentation methods to assess the impact of our enhanced images on downstream tasks
SSD~\citep{liu2016ssd}, Refinedet~\citep{zhang2018refinedet} and PAA~\citep{kim2020paa} are used as comparison methods for object detection tasks. LEDNet~\citep{wang2019lednet} and CGNet~\citep{wu2020cgnet} are employed to evaluating the performance on semantic segmentation tasks.
For a fair comparison, we used the same detection and segmentation networks and trained them with results obtained from different enhancement methods using identical training parameters. The results of all methods on the RUIE detection dataset are shown in Fig.~\ref{fig:UCCSdetection}, where our method not only achieves superior enhancement but also outperforms other methods in the detection of distant urchins and trepangs. Fig.~\ref{fig:aquarium} shows the results on the Aquarium dataset, where the presence of ripples on the water surface poses challenges for network inference. It is evident that all enhancement methods are affected by ripples, leading to multiple detections, while our method is less affected. Additionally, our method distinguishes fish and sharks more clearly, avoiding detection errors. PR and ROC curves for the detection results are also plotted, as shown in Fig.~\ref{fig:detetion_duibi2}. Clearly, the proposed method achieves the best results on both RUIE and Aquarium datasets. For semantic segmentation tasks, the results on the SUIM dataset for all methods are presented in Fig.~\ref{fig:suim}, highlighting that our method significantly reduces segmentation errors. Detailed quantitative evaluations based on IoU metrics are displayed in Fig.~\ref{fig:segmentation_duibi2}. Our method outperforms all other methods across almost all categories. Therefore, based on both quantitative and qualitative experiments, our method's implicit extraction of semantic information during enhancement leads to superior performance in subsequent perceptual tasks.

\begin{figure*}[!htb]
	\centering
	\setlength{\tabcolsep}{1pt}
	\includegraphics[width=0.99\textwidth]{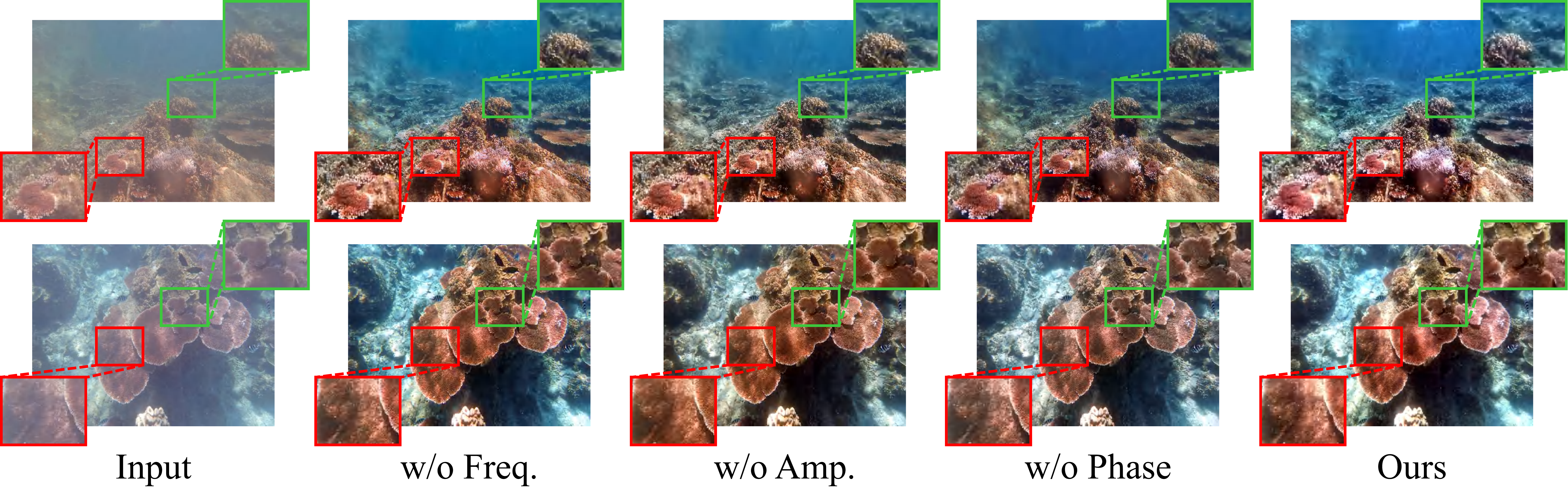}
	\caption{Visualization for ablation study on the SFA. w/o Freq. represents that the proposed method does not consider both amplitude and phase components.}
	\label{fig:visualizationforfre}
\end{figure*}

\begin{figure}[!htb]
	\centering
	\setlength{\tabcolsep}{1pt}
	\includegraphics[width=0.490\textwidth]{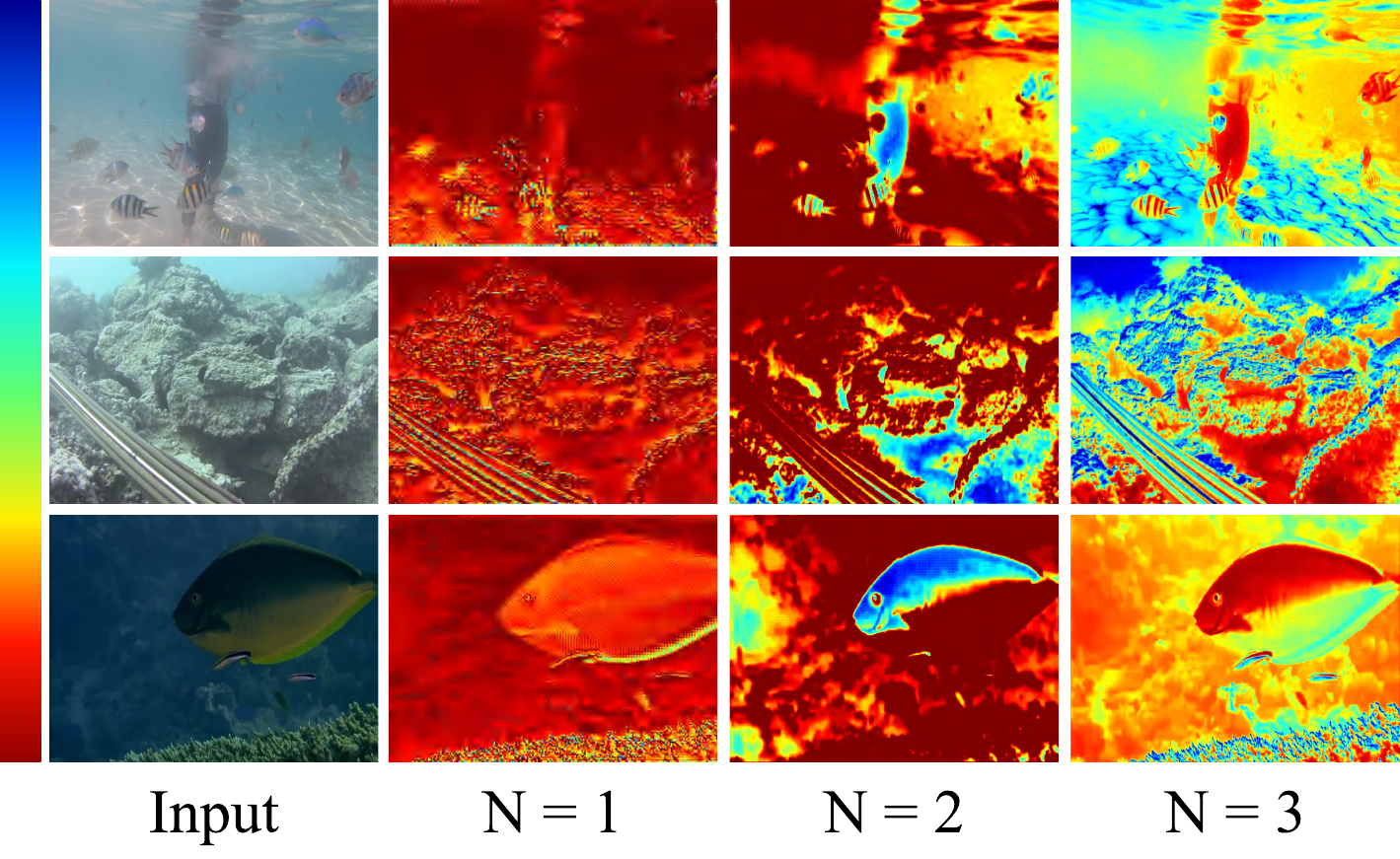}
	\vspace{-10pt}
	\caption{Visualization of image translation process in the proposed Heuristic Invertible Network.}
	\label{fig:visualizationforblocksheatmap}
\end{figure}

\begin{figure}[!htb]
	\centering
	\setlength{\tabcolsep}{1pt}
	\includegraphics[width=0.499\textwidth]{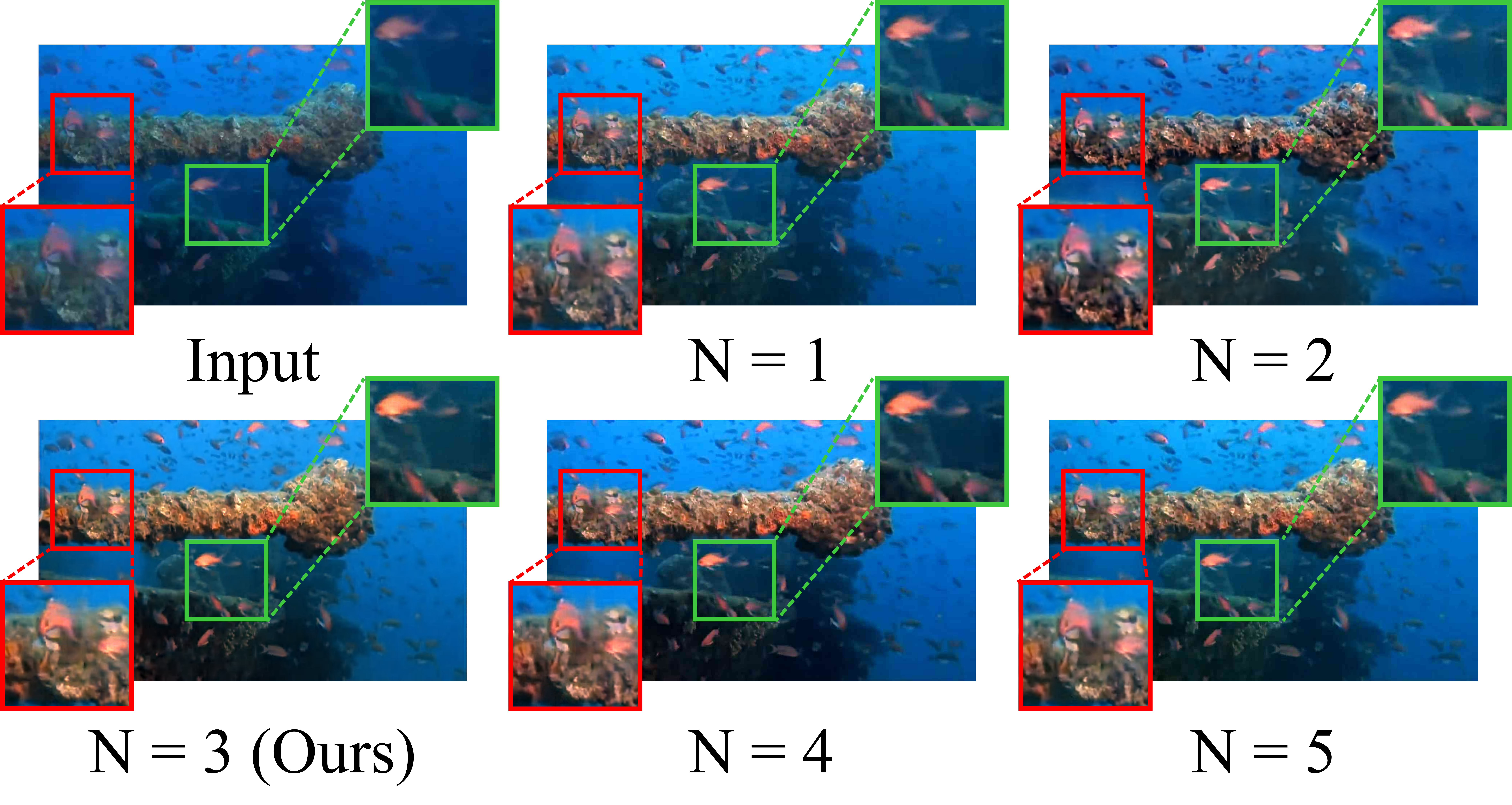}
	%	\vspace{-18pt}
	\caption{Visualization for ablation study on the number of Hybrid Invertirble Blocks.}
	\label{fig:visualizationforblocks}
\end{figure}
\begin{figure}[!htb]
	\centering
	\setlength{\tabcolsep}{1pt}
	\includegraphics[width=0.499\textwidth]{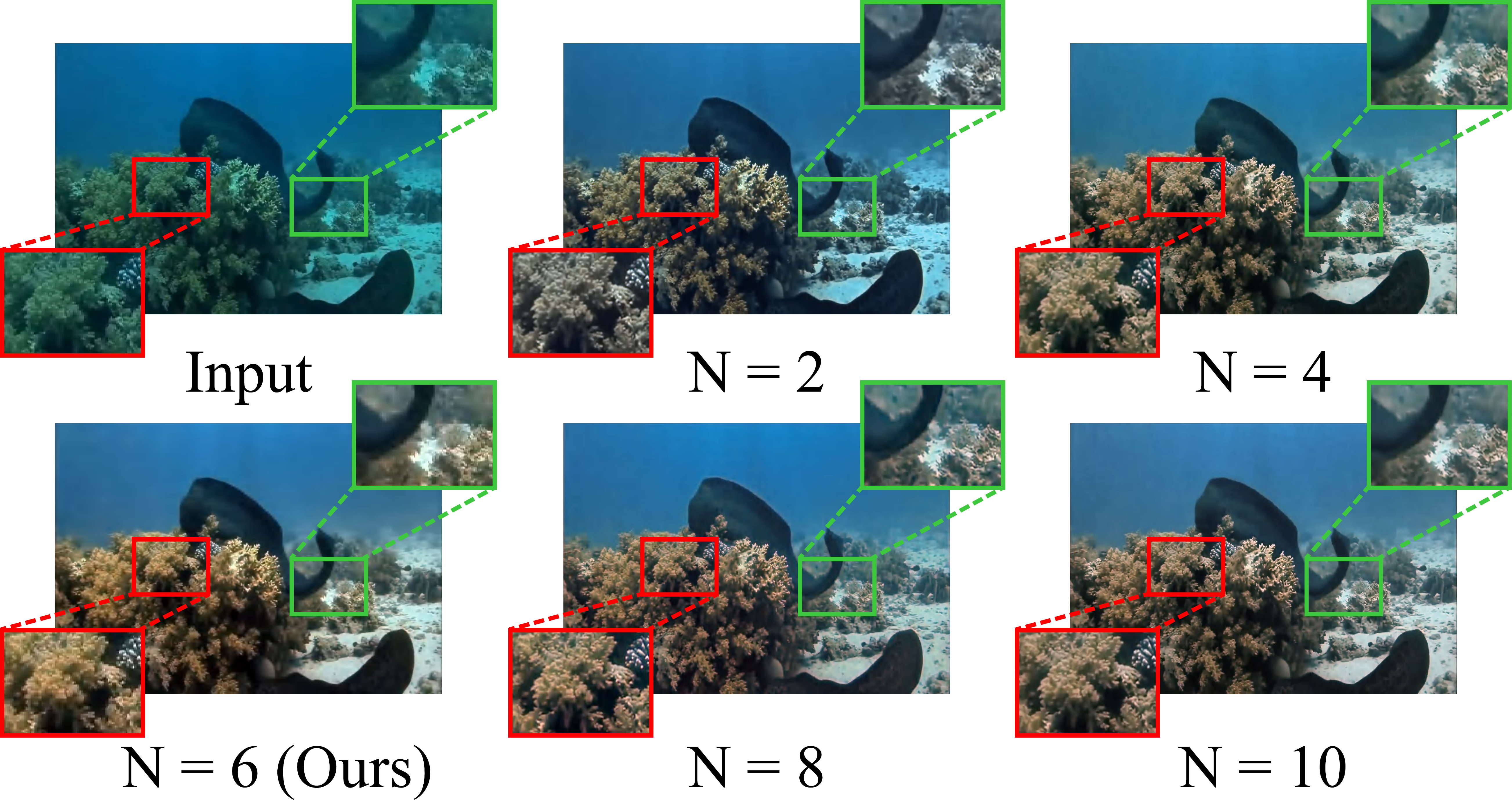}
%	\vspace{-18pt}
	\caption{Visualization for ablation study on the number of flow steps.}
	\label{fig:visualizationforflows}
\end{figure}

\begin{figure*}[!htb]
	\centering
	\setlength{\tabcolsep}{1pt}
	\includegraphics[width=0.98\textwidth]{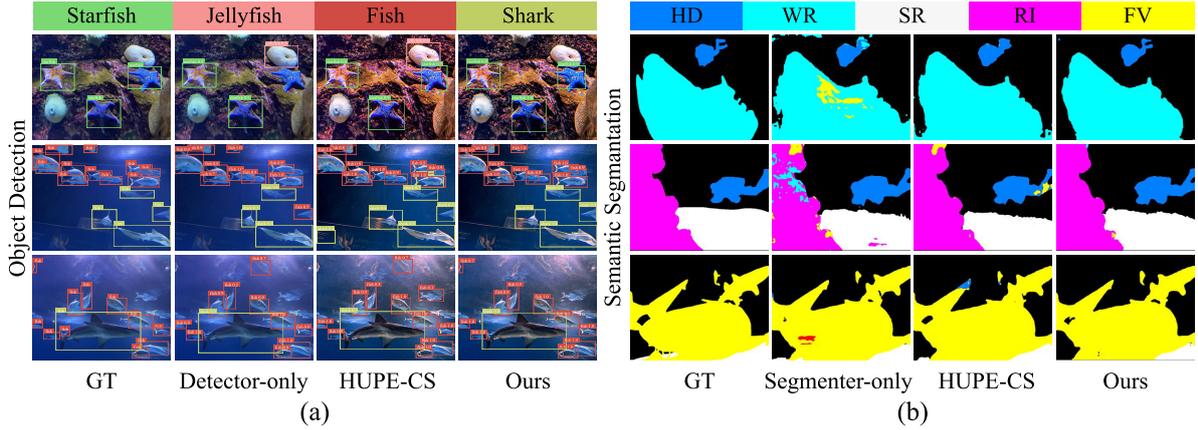}
%	\vspace{-3pt}
	\caption{Visualization results for underwater image enhancement in object detection and semantic segmentation. }
	\label{fig:ablation_TPM_vis}
\end{figure*}
\begin{figure*}[!htb]
	\centering
	\setlength{\tabcolsep}{1pt}
	\includegraphics[width = 0.95\textwidth]{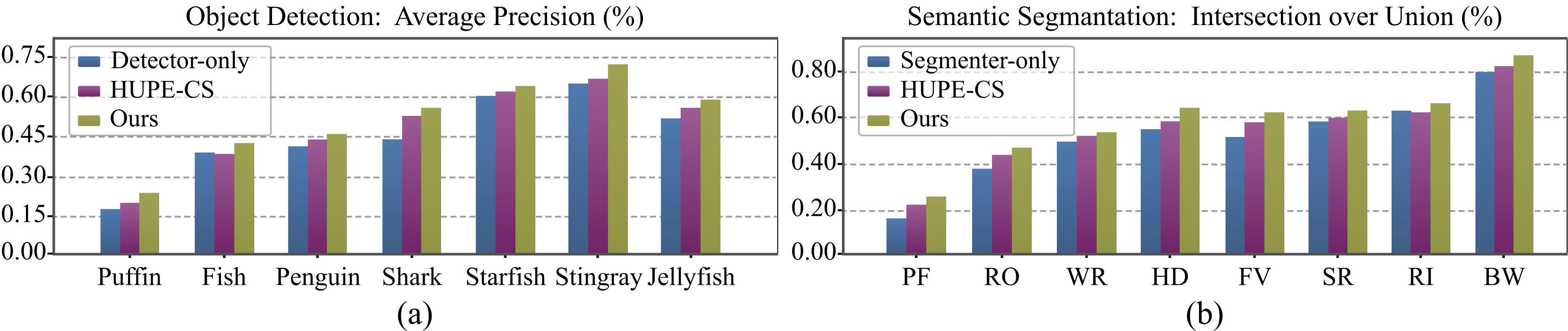}
	%	\vspace{-8pt}
	\caption{Quantitative comparison for underwater image enhancement in object detection and semantic segmentation. We use Average Precision (AP) to evaluation the detection task and Intersection over Union (IoU) to evaluation the segmentation task.}
	\label{fig:ablation_TPM}
\end{figure*}

\subsection{Ablation Study}
\subsubsection{Study on Loss Function}
We verify the effectiveness of all loss functions. The visual effect is shown in Fig.~\ref{fig:visualizationforlossablation}. 
In the first example, when the $\mathcal{L}_\mathrm{s}$ loss is ignored during training, the contrast enhancement effect is limited. When $\mathcal{L}_\mathrm{1}$ and $\mathcal{L}_\mathrm{c}$ losses are not used, excessive enhancement of the red channel causes the background water plants to deviate from their own colors. In the second example, compared to other ablation methods, only the proposed method simultaneously restores the color and brightness of the foreground and background, producing a visually pleasing image, especially in the content of the green box. The quantitative results are shown in (a) of Fig.~\ref{fig:lossablation}, which confirms the effectiveness of all loss functions during the training process.
\subsubsection{Study on Heuristic Prior guided Encoder}
We verify the effectiveness of the input of the proposed Heuristic Prior guided Encoder. The visualization results are shown in Fig.~\ref{fig:visualizationforinputencoder}. Both scenes have uneven blue artifacts. 
When the input does not have RGB images, the enhancement effect produces relatively dark results.
Although the original color of the image can still be restored well when the input has no gradient and depth. However, after adding the gradient and depth maps, the brightness and contrast of the enhanced image are significantly improved. The quantitative results of the ablation experiment are shown in (b) of Fig.~\ref{fig:lossablation}, and it is clear to see the effectiveness of the proposed HPE.

\subsubsection{Study on Spatial-Frequency Affine Block}
In order to verify the effectiveness of the proposed Spatial-Frequency Affine Block, we neglected the amplitude branch and the phase branch respectively for the ablation study. The visualization results are shown in Fig.~\ref{fig:visualizationforfre}. 
When both phase and amplitude components are not introduced in the enhancement process, the enhancement images still suffer from obvious color cast in the distance.
When phase information is not used, the enhanced image fails to recover bright scene reflections. 
While neglecting amplitude information still successfully restores colors underwater, the overall brightness of the image remains lower than that of the final version which considers both phase and amplitude information, especially in the red frames. The quantitative results of the experiment are shown in (c) of Fig.~\ref{fig:lossablation}, consistent with the visualization results.
\subsubsection{Study on Number of Hybrid Invertible Blocks}
Increasing the number of Hybrid Invertible Blocks (HIBs) can enhance the network's depth to facilitate the transformation of underwater images into their clear counterparts. 
Fig.~\ref{fig:visualizationforblocksheatmap} shows a visualization of the augmentation process of our method. N represents the N-th Hybrid Invertible Block. It can be clearly seen that the input image gradually improves the contrast and clarity of features during the restoration process, achieving asymptotic image enhancement.
%, as visualized in Fig.~\ref{fig:visualizationforblocksheatmap}. 
However, higher depth also demands more computational resources. Therefore, we conducted an ablation study on the number of HIBs, and the visualization of this experiment is presented in Fig.~\ref{fig:visualizationforblocks}. It is evident that when $N$ is less than 4, a greater number of HIBs leads to more pronounced color restoration. However, beyond $N>3$, the color correction effect starts to weaken. The quantitative results of the experiment are shown in (c) of Fig.~\ref{fig:lossablation}, consistent with the visualization results. When the number os HIBs are less than 3, all evaluation metrics are significantly lower, indicating that fewer HIBs are insufficient for effective underwater image restoration. On the other hand, when $N$ is increased, adding more HIB modules does not further improve image quality. Therefore, considering both quantitative and qualitative results, we ultimately trained the network with 3 HIBs.
\subsubsection{Study on Number of Flow Steps}
In the structure of each Hybrid Invertible Block, multiple flow steps are employed. To strike a balance between computational resources and enhancement effectiveness, we conducted ablation experiments on the number of flow steps. Visual results are shown in Fig.~\ref{fig:visualizationforflows}, where the underwater scene exhibits a noticeable blue tint due to the attenuation of red and green light. When $N$ is less than 4, the enhanced images still suffer from significant impact, as indicated by the red box. The seaweed next to the black fish appears overall white, deviating from its original color. When $N$ is set greater the 4, the enhanced image demonstrates balanced color distribution and clear details, especially when $N$ is equal to 6, which is closer to natural in red frame. Quantitative results in (d) of Fig.~\ref{fig:lossablation} show that the quantitative results are consistent with the qualitative ones. When $N$ is greater than or equal to 6, the quantitative results for the enhancement are very close. Considering computational resources, we ultimately set 6 Flow Steps in each HIB.
%\subsubsection{Study on Subsequent Perception Tasks}
\subsubsection{Study on Semantic Collaborative Learning Module}
To validate the effectiveness of our method for downstream task applications, we first trained and tested the underwater original images using the same detector and segmenter. Subsequently, 
%without incorporating the Task Perception Module, 
we trained our network with the traditional cascade strategy~\cite{zhang2023waterflow} without the proposed semantic collaborative learning module~(HUPE-CS).
%and separately trained the subsquent network on the enhanced images. 
The visual results obtained are shown in (a) of Fig.~\ref{fig:ablation_TPM_vis}, and for both examples, it is evident that the proposed enhancement network, designed to avoid introducing artifacts and improve network robustness, continues to benefit the enhancement network even without SCL. When the proposed SCL is combined in the training process, the detection performance is further enhanced. Similarly, for semantic segmentation experiments, the visual results in (b) of Fig.~\ref{fig:ablation_TPM_vis} align with the results of the detection task. Quantitative results are presented in Fig.~\ref{fig:ablation_TPM}, demonstrating that both our network and the proposed training strategy further enhance adaptability to downstream tasks.
\section{CONCLUSION}
In this paper, we introduce a downstream task-driven heuristic invertible network for underwater perceptual enhancement. We first adopt the invertible network into underwater image enhancement process to establish the invertible translation with bilateral constraints between the underwater image and its clear counterpart. Then, the heuristic prior guided injector is embedded into the data-driven invertible mapping to help the enhancement network better characterize underwater scene characteristics by progressively adopting different modalities of knowledge.
At last, we introduce the semantic collaborative learning module, which propagates high-level perceptual features into the enhancement model to help generate the task-driven enhanced image. Extensive experiments on multiple benchmarks show that the proposed method not only achieves good visual enhancement effects, but is also more suitable for perception tasks.
\section*{Acknowledgment}
This work was supported in part by the National Natural Science Foundation of China (Nos. U22B2052, 12326605, 62027826, 62302078, 62372080); 
in part by National Key Research and Development Program of China (No. 2022YFA1004101);
and in part by China Postdoctoral Science Foundation (No. 2023M730741).
%This work is partially supported by the National Natural Science Foundation of China under Grant U22B2052, 62302078, 62372080, 62027826 and 12326605; and in part by China Postdoctoral Science Foundation under Grant 2023M730741.


\bibliography{sn-bibliography}% common bib file

\end{document}